\documentclass[Numbered,pdflatex,sn-basic,iicol]{sn-jnl}

\usepackage{graphicx}
\usepackage{multirow}
\usepackage{amsmath,amssymb,amsfonts}
\usepackage[title]{appendix}
\usepackage{xcolor}
\usepackage{booktabs}

\begin{document}

\title[Rethinking localization on chest radiographs]{Rethinking weakly
supervised localization on chest radiographs: position nulls, protocol
dependence, and PACE}

\author*[1]{\fnm{Shixin} \sur{Qiu}}\email{qiushixin8282@gmail.com}

\affil*[1]{\orgdiv{School of Information Science and Technology},
           \orgname{Sanda University},
           \orgaddress{\city{Shanghai}, \country{China}}}

\abstract{Weakly supervised localization on chest radiographs is reported as a
single overlap accuracy, and those numbers are not comparable across papers. We
measure what one contains. Against position-null controls that keep a readout's
own box and move only its position, class activation mapping and its
gradient-weighted variant beat the null on two to four of seven annotated
findings on ChestX-ray14 and two of eight on CheXlocalize. The null is beatable:
three board-certified radiologists clear it on four to seven. Three rarely stated choices each move a result by more than most reported
method effects: the results regime, the extraction threshold, and the mask-to-box
rule, whose cost is predicted by how fragmented the thresholded mask is.
Predicted box area alone reverses a paired comparison. These measurements produced PACE, for \emph{population anomaly and
class evidence}, which multiplies the finding identity from the label-trained
head by the spatial evidence from a frozen radiograph backbone that never sees a
pathology label. It is the case study that made the
protocol dependence visible, and we report its costs beside its gain: in an
exploratory matched-area analysis it improves strict-overlap accuracy on both
corpora and beats a label-derived second map, but it is the least faithful of the three readouts to
the classifier's causal map, its advantage depends on the box rule, and, this
benchmark releasing no localization validation set, every design choice behind it
was made on the boxes it is scored on. Its one clean transfer beats one baseline
and loses to the other.}


\maketitle

\section{Introduction}\label{sec:intro}

A chest radiograph model trained on hospital-scale data learns from image-level
labels: a study is recorded as showing an effusion, with nothing said about
where. Clinical use asks for more than the label. A reader who is shown a
probability and no location cannot check the model against the image, and the
weakly supervised localization literature exists to close that gap by reading a
spatial map out of a classifier that was never given a box
\cite{bib-zhou2016,bib-selvaraju2020,bib-wsolsurvey}.

The standard readouts take that map from the classifier itself. Class activation
mapping projects the classifier weights back onto the last feature map, and
gradient-weighted mapping replaces those weights with pooled gradients
\cite{bib-zhou2016,bib-selvaraju2020}. Both have been applied to chest
radiographs many times, and later work improves them by adding a second spatial
map: a learned class-specific foreground mask constrained by anatomical priors
\cite{bib-clarity}, a per-class occurrence prior averaged from training-set maps
\cite{bib-jung2023}, or a prototype layer whose prototypes are projected onto
training patches of the matching class \cite{bib-cipl}.

The mechanism those methods share is also their limitation, and it is a property
of the data rather than of the architecture. ChestX-ray14's labels are not read
from the images. They are mined by a natural language pipeline from free-text
reports, and the dataset paper states that no gold-standard labels exist for the
collection \cite{bib-wang2017}. A radiologist review of around 700 images from
this dataset and one other found that its labels do not accurately reflect the
visual content, with positive predictive values mostly 10 to 30 points
\emph{below} the values in its own documentation, alongside hidden stratification
and label disambiguation failures \cite{bib-oakdenrayner2020}; separately, a gold
standard assembled for a different corpus and a different miner found that report
mining can overstate a finding by a factor of several \cite{bib-cxrlt}. The first
is evidence about these labels and the second about the mechanism, and we keep
them apart. Every second map listed above is built from
those image-level labels, so a map that answers \emph{where} is derived from the
same annotation process that answers \emph{whether}, and inherits its errors. A
representation that has never seen a pathology label cannot inherit them
\emph{through that supervision}, though it remains exposed to correlated
acquisition, population and pretraining biases. That is the motivation for what
follows, and we are explicit that it is not among the
results: we do not corrupt labels and compare degradation curves, so this paper
shows that the two streams carry complementary evidence and does not show that
either is the more robust to report-mining error.

That observation is a hypothesis, not a result. We therefore begin with two
measurements. The first tests how much of a localization score on this benchmark
is retained by controls that keep the model's own predicted box and change only
where it sits; the practice is
imported from natural-image evaluation \cite{bib-choe2020}, where it is standard,
and no chest radiograph study we are aware of reports it. The second asks whether
the pathology labels are the source of the positional signal, by building a
readout from a frozen backbone that never saw one. The answers are that a large
part of a macro score is positional, that one class is satisfied by a fixed box,
and that the label-free readout lands in the same accuracy range as class
activation mapping.

The method follows from those two answers rather than preceding them. We call it
PACE, after the two quantities it multiplies: the \emph{population anomaly} of a
patch and the \emph{class evidence} for a finding. It takes the finding identity
from the label-trained head, takes the spatial evidence from a frozen radiograph
backbone that scores each patch against the same patch position across a
reference population, and multiplies the two normalised maps.
The product is chosen against four alternative combination rules, and the only
rule that scores below both single streams is the one that fires when either
stream is confident, so the two behave as a conjunction: a finding must be
evidence for the class and unusual against the population. PACE adds no training
of any kind; it is an inference-time readout over a frozen classifier.

The measurements are the primary claim, and the reason is that they apply a
diagnostic imported unchanged from natural-image evaluation \cite{bib-choe2020}
to readouts this paper did not design, so they inherit no development history of
its own. We quantify the position prior in two corpora, establish a human ceiling
for the same test, and show that the results regime, extraction threshold and
mask-to-box rule can each move a score more than a reported method effect. We then compare overlap with occlusion-based
faithfulness. PACE is a case study produced by those measurements: it combines
a label-trained class map with a label-free population-anomaly map, without
additional training. Its overlap gain comes with two costs. It is the least
faithful readout tested, and its advantage depends on the mask-to-box rule.
Because ChestX-ray14 has no localization validation set, we also report a
patient-level post-hoc stability split and a frozen transfer to CheXlocalize.
The former tests stability, not confirmation; the latter beats one baseline and
loses to the other.

\section{Related work}\label{sec:related}

ChestX-ray14 established the benchmark: 112,120 frontal radiographs of 30,805
patients, fourteen findings mined from reports, and a patient-level official
split \cite{bib-wang2017}. CheXpert followed with 224,316 radiographs, grouped into
187,641 training studies, and an explicit
uncertain state on every observation \cite{bib-irvin2019}, which needs a stated
policy rather than a default; the options were surveyed by its authors and
extended by later work that maps the uncertain state to an interval instead of a
point \cite{bib-pham2021}.

Reported results on ChestX-ray14 fall into two bodies that do not cite each
other. One runs from the dataset paper's own 0.745 through a plain DenseNet-121
at 0.805 to 0.807 \cite{bib-ranjan2018,bib-wangh2021} to the strongest recent
official-split method at 0.828 \cite{bib-cipl}. A second reports 0.849 to 0.870
while quoting the same split sizes and citing none of the first
\cite{bib-synthensemble,bib-graphguided2026,bib-cbamcxr}. The second body starts
0.021 above where the first ends, so a number drawn from it is not comparable with
one drawn from the first and the difference is not a method effect. Everything below belongs to the first. A widely quoted 0.8414
\cite{bib-chexnet} is measured on a random split and is excluded for that
reason; the same architecture appears at 0.814 in an official-split table
\cite{bib-cipl}.

Class imbalance on this benchmark is severe and has its own literature, from
reweighting by an effective sample count \cite{bib-cui2019} to losses designed
for the long tail of multi-label recognition \cite{bib-lin2023} and to the
challenge series that made the tail its subject \cite{bib-cxrlt}. We use this
literature as a diagnostic in Section~\ref{subsec:analysis} rather than as a
contribution.

\noindent\textbf{Localization readouts.}\quad Class activation mapping recovers a spatial map by projecting classifier weights
onto the final feature map \cite{bib-zhou2016}, and gradient-weighted mapping
generalises it to architectures without a global pooling head
\cite{bib-selvaraju2020}. Both remain the default readouts on chest radiographs, whether applied directly
or through a self-supervised attention stage \cite{bib-dsouza2023}.
The natural-image community has been more sceptical of the resulting numbers than
the medical one: a controlled re-evaluation showed that much of the reported
progress in weakly supervised localization did not survive a fair protocol, and
that a trivially positioned box is a strong baseline \cite{bib-choe2020}. That
critique supplies the positional controls used here. It has not, to our
knowledge, been applied to a chest radiograph benchmark, where a further
difficulty applies: the 984 released boxes were each drawn by one radiologist who
was shown the mined keyword first \cite{bib-wang2017}.

A separate criticism attacks overlap itself. Agreement with a radiologist box
cannot tell whether a map has anything to do with the prediction, so a map can
sit on the right anatomy for the wrong reason; measured by occluding patches and
correlating the resulting score changes against the map, medical vision-language
attention comes out anti-correlated on six of the seven probes that study reports
\cite{bib-grounding}. Its own positive controls are a DenseNet-121 and a
ResNet-50 read out with gradient-weighted mapping, which are two of the three
architectures used here, so the criticism as published does not extend to
class-activation readouts on chest radiographs and the open question is what it
says about a readout that mixes in a second stream. That is what
Section~\ref{subsec:faithful} measures.

\noindent\textbf{Fusing a second spatial map.}\quad The line closest to this work multiplies the class-conditional map by a second
spatial map at inference. CLARiTy multiplies its class attention map
by a learned class-specific foreground mask, with anatomical priors constraining
that mask through training losses \cite{bib-clarity}. Jung et al.\ average
training-set class activation maps into a per-class occurrence prior and retrain
under it \cite{bib-jung2023}. Two nearby methods are outside the line: one forms
a class-agnostic foreground map by averaging maps over all categories and
thresholds it directly \cite{bib-fpm}, and one emits a scalar image-level anomaly
score rather than a spatial map \cite{bib-capcam}.

The second map in both members is built from the image-level pathology labels.
CLARiTy's mask is class-specific and learned under them; the occurrence prior is
averaged from maps produced by a network trained on them. PACE's second map has
never seen one, and that source difference is its entire novelty claim.

\noindent\textbf{Prototype-based interpretation.}\quad CIPL reports the strongest result in the official-split literature and is the
second closest prior work \cite{bib-cipl}. We quote its classification number,
do not reproduce it, and do not compare against its localization at all. The
reason is stated here rather than left to be inferred from a table entry: its
localization model is trained on all 111,240 images without box annotations,
which by a filename join against our frozen manifests contains 24,716 images of
the official test split, so its localization protocol and ours are not the same
protocol and a shared table would be misleading in our favour or in theirs
depending on the row. It scores each position by similarity to
class-specific prototypes and, after every epoch, projects each prototype onto
the nearest training patch feature of the corresponding class, so the prototypes
are selected using the image-level labels; the localization mask is then the
averaged class activation map thresholded at 0.5. Its chest radiograph backbone
is DenseNet-121 and its box subset is the same released one, so architecture and
benchmark are shared and the difference lies in where the spatial evidence comes
from, exactly as against CLARiTy.

Two qualifications narrow this comparison. CIPL's backbone is initialised from ImageNet
\cite{bib-imagenet} and its intra-image alignment term needs no labels, so parts
of its spatial signal are label-free too; and PACE is not label-free either,
since its output multiplies a map read from a label-trained head. The supportable
statement is narrower than either extreme: CIPL's \emph{second} source of spatial
evidence is chosen by the pathology labels and PACE's is not, while the ImageNet
initialisation is common to both and is therefore no difference at all.

CIPL also argues that post-hoc saliency is unreliable unless additional
regularisation objectives constrain it during training. PACE is post-hoc and
unregularised, so this criticism applies directly. The measurement in
Section~\ref{subsec:faithful} supports PACE on localization overlap but not on
faithfulness to the classifier.

\noindent\textbf{Open measurement gaps.}\quad Three gaps remain. First, no chest radiograph localization study we have found
reports a positional control, so published macros mix grounding with a position
prior that is never separated out. Second, every published second-map fusion on
this benchmark derives its second map from the same automatically mined labels
whose errors it would need to be independent of. Third, localization claims are
almost always made on overlap alone, which cannot distinguish a map that
describes the prediction from one that merely lands on plausible anatomy.

\section{Data, metrics, and the position-null instrument}\label{sec:instrument}

\subsection{Datasets, metrics and implementation}\label{subsec:data}

\textbf{ChestX-ray14} \cite{bib-wang2017} supplies 112,120 frontal radiographs
from 30,805 patients with fourteen findings. We use the official patient-level
split and carve a validation set out of its training pool, giving
77,919 / 8,605 / 25,596 images. We verified the test split image by image
and label by label against two independent third-party copies: the split files
released with a masked-autoencoder study \cite{bib-xiao2023} and those released
with a foundation-model study \cite{bib-evax}: the same 25,596 images, zero label
mismatches on all fourteen classes, and no test image in our training or
validation sets. Localization uses the released box file, 984 boxes over 880
images covering 8 of the 14 findings; all 880 lie in the official test split,
which we confirmed by joining the box file against our frozen manifests.

\textbf{CheXpert} \cite{bib-irvin2019} supplies the second dataset. Our copy
reproduces its published training-set size of 223,414 images exactly, and all 28
label counts of its Table~1 reproduce when counted per study rather than per
image, which is the denominator that source uses. We carve it frontal-only at
patient level into 152,550 / 9,425 / 29,052. Its uncertain state
is treated as an ablation axis and not as a setting; the default masks uncertain
entries as unobserved rather than asserting a value for them. \texttt{No Finding}
is dropped, so CheXpert is scored over 13 classes and its mean is not comparable
with papers that score 14. CheXpert ships no boxes, so it carries classification
and cross-dataset transfer only; the asymmetry is a property of the data.

\textbf{Results regime.} As set out in Section~\ref{sec:related}, this
benchmark carries two non-communicating bodies of classification results. Every
number in this paper belongs to the official-split regime, and no number from the
other appears in any table here. Appendix~\ref{subsec:repro} reproduces a
published baseline matched on architecture and split, and
Appendix~\ref{app:compare} scores eight released checkpoints on the same frozen
manifest with the same metric code rather than quoting their published values,
which is what validates the evaluation code every later row depends on.

\noindent\textbf{Evaluation metrics.}\label{subsec:metrics}\quad Classification is the mean over classes of the area under the receiver operating
characteristic curve. The crop protocol is stated with every value, because
ten-crop testing inflates against single-crop on this benchmark by an amount
comparable to the differences being discussed.

A localization outcome is binary per box: the predicted box either reaches a
threshold $T$ against the annotation or it does not, over
$T \in \{0.1, 0.25, 0.5, 0.75\}$ and under two overlap measures. Intersection
over union divides by the union; intersection over the predicted box divides by
the predicted area alone.

\emph{Two aggregates of those outcomes appear in this paper and they are
different numbers.} The \textbf{pooled accuracy} is the fraction of all 984 boxes
that clear $T$; the \textbf{class macro} is the mean over the eight annotated
findings of the same fraction computed within a class. They differ because the
classes carry 79 to 180 boxes each, so pooling weights Atelectasis more than
twice Nodule. Every paired comparison \emph{between two readouts} is pooled, because the
pairing is per box, so Table~\ref{tab:main} and the $p$-values and intervals that
accompany it are pooled quantities. Every number compared against a per-class
control or reported per class is a macro: Table~\ref{tab:prior},
Table~\ref{tab:ablation}, Figure~\ref{fig:prior} and Figure~\ref{fig:rule}. Two
statistics are on neither footing and are labelled where they appear: the
per-class McNemar tests of Section~\ref{subsec:rq1} are computed within one class,
so their sample is that class's 79 to 180 boxes and not the 984, and the intervals
beside a position-null margin come from a bootstrap stratified by class, which
resamples a macro. Each table states which aggregate it carries, and no difference
of macros is ever quoted as a paired difference.

We use intersection over union as the \textbf{protected metric}, meaning the one
every claim in this paper is required to survive, and report intersection over
the predicted box beside it. The reason is that the second divides by the
predicted area alone, so a readout that emits a smaller box scores higher on it
with no change in where the box sits, and the readouts compared here emit boxes
of markedly different sizes: on DenseNet-121 the median predicted-to-annotated area
ratio is 1.40 for PACE, 2.80 for gradient-weighted mapping and 5.92 for class
activation mapping, against an annotation that covers 8.2 percent of the image.

Intersection over union is not neutral to box area either, and we state the
direction rather than leave it implied. It penalises an oversized box as hard as
an undersized one, so a readout whose box is $\rho$ times the annotation cannot
exceed $1/\rho$ however well centred: 0.17 at $\rho = 5.92$ and 0.71 at
$\rho = 1.40$. What it removes is the free ride, since shrinking a box that is
not on the finding lowers it. Two consequences are carried through the paper. A
\emph{between-readout} statistic matched to the predicted area, which includes
intersection over the predicted box and the margin over a size-matched control,
is not used as evidence for PACE, because the sizes differ. A
\emph{within-readout} comparison against that readout's own size-matched control
is unaffected by this, since both boxes then have the same area by construction,
and Section~\ref{subsec:rq1} therefore reports it for PACE as well as for the two
baselines. The residual threat, that PACE's advantage at loose thresholds is
concentration rather than relocation, is stated as an open one in
Section~\ref{sec:conclusion}.

\noindent\textbf{Box extraction, one rule for every readout.}\quad Every map
compared here, both baselines, each single stream and every combination rule, is
turned into a box by exactly the procedure of Section~\ref{sec:method}, and
$\tau$ is reported with every number because results at different $\tau$ are not
comparable. The class-evidence stream lives on a $7 \times 7$ grid and the frozen
transformer of the population-anomaly stream on $14 \times 14$; the latter is
average-pooled to $7 \times 7$ before combination, so a finer map cannot deliver a
tighter box for a reason unrelated to labels.

\textbf{Uncertainty.} Every headline number is the mean and standard deviation
over three seeds. For localization the seed spread is not the uncertainty that
matters: each class carries only 79 to 180 boxes, so the binomial standard
deviation is 0.02 to 0.054, comparable to or larger than the spread over seeds.
The 984 boxes are not 984 independent units: they fall on 880 images and 93 of
those carry more than one box, which share the input, the classifier output and
the heat map. We therefore report every significant comparison again under a
bootstrap that resamples \emph{images} rather than boxes
(\texttt{scripts/cluster\_robust.py}), and say so with the result.
Because a model box and a size-matched control box differ only in position, and
because two readouts are scored on the same boxes, the comparisons are paired, so
per class we use an exact McNemar test \cite{bib-mcnemar1947}, across classes a
Holm correction \cite{bib-holm1979}, and for the macro a stratified bootstrap
\cite{bib-efron1979} that reselects the strongest control inside every resample,
so that the interval carries the control-selection uncertainty. Across seeds, the governing $p$-value
depends on the direction of the claim: for a claim that one readout beats
another we report the largest $p$ over seeds, so that significance in the worst
seed implies it in all; a claim that two are indistinguishable would instead be
governed by the smallest, and we do not make one from a failure to reject.

\noindent\textbf{Baselines and implementation.}\quad Classifiers are DenseNet-121
\cite{bib-densenet}, ResNet-50 \cite{bib-resnet} and ConvNeXt-Tiny
\cite{bib-convnext}, each initialised from ImageNet \cite{bib-imagenet}. The
frozen backbone for the population-anomaly stream is the base variant of a
masked-image-modelling model pretrained on 520k radiographs \cite{bib-evax};
two alternatives are reported for contrast, a masked autoencoder \cite{bib-mae}
pretrained on 266k radiographs \cite{bib-xiao2023} and a self-supervised vision
transformer \cite{bib-vit} pretrained on 142M natural images \cite{bib-dinov2}.
Each frozen backbone receives its own pretraining normalisation, read from that
project's own transform code rather than assumed. The reference bank holds
$N = 1000$ unlabelled radiographs drawn from the training pool, which contains
no box-annotated image; the test-split draw, from which they are excluded, is
the transductive control of Section~\ref{subsec:main}. Appendix~\ref{app:impl}
gives the hardware, the software versions and the full training recipe.

\section{What a localization score contains}\label{subsec:rq1}

\begin{figure*}[!tp]
\centering
\includegraphics[width=\textwidth]{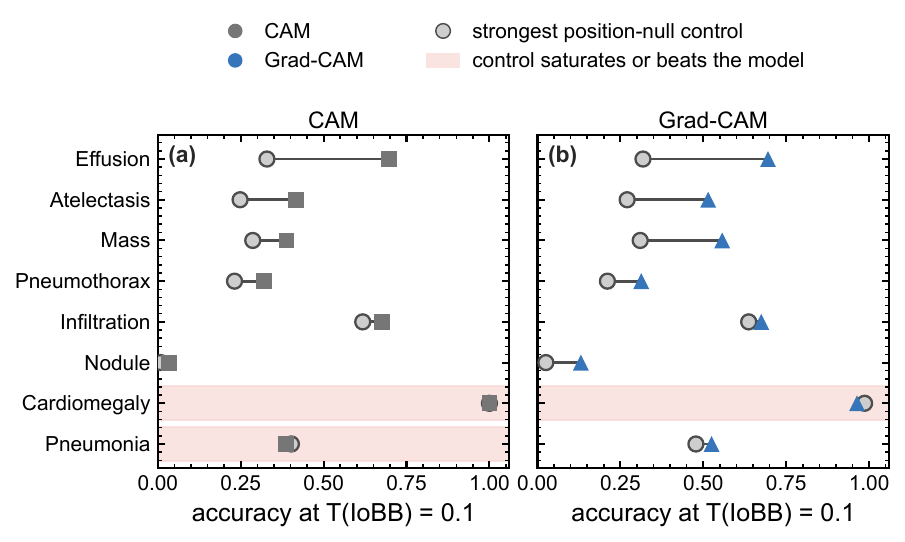}
\caption{How much of a localization score on this benchmark is a position prior.
Per class, the accuracy of (a) class activation mapping and (b)
gradient-weighted mapping on DenseNet-121 at three seeds, against the strongest
of three position-null controls. Each control reuses the model's own predicted
box dimensions and changes only its position, so the pair is size-matched and the
gap measures grounding rather than box area. Shading marks a class whose control
saturates or beats the model. Cardiomegaly's control scores exactly 1.0000 under
class activation mapping, so that class is satisfied by a fixed box, carries no
localization information at this threshold, and is excluded from the informative
macro; under the same readout the control also beats the model on Pneumonia.
Classes are ordered by the margin in panel (a) and the order is held across both
panels. The practice of scoring against trivial position baselines is taken from
\cite{bib-choe2020}.}
\label{fig:prior}
\end{figure*}

Every readout is scored against three controls that reuse the model's own
predicted box dimensions and change only its position: a translated box, a
uniformly random box, and a box fixed at the image centre. The strongest of the
three is taken per class, which is the conservative choice. The practice is
imported from \cite{bib-choe2020}; the controls null position and not size, so
they are not model-free baselines and what they test is the positional claim
alone. Figure~\ref{fig:prior} gives the per-class picture and
Table~\ref{tab:prior} the summary.

\begin{table*}[!tp]
\caption{Position-null controls on DenseNet-121, three seeds, 984 boxes, class
macros. Each readout is scored against \emph{its own} controls, which reuse that
readout's predicted box size and change only its position, so every column is a
within-readout test and the box-size confound of Section~\ref{subsec:metrics}
does not enter it. Intersection over the predicted box is sound here and only
here, for the same reason, and the protected metric is given below it. Informative
classes exclude Cardiomegaly, whose control is saturated. Classes won are those
significant in every seed after a Holm correction across the seven informative
classes; the count, not the margin, is what compares across columns, because the
margins are measured against controls of three different sizes. PACE is included
because a diagnostic this paper introduces and then applies only to the methods it
competes with is not a diagnostic.}
\label{tab:prior}
\begin{tabular}{lccc}
\toprule
 & Class activation & Gradient-weighted & PACE \\
 & mapping & mapping & \\
\midrule
\multicolumn{4}{l}{\emph{Intersection over the predicted box, $T = 0.1$}} \\
Macro, all 8 classes        & 0.4896 & 0.5467 & 0.6098 \\
Strongest control           & 0.3918 & 0.4057 & 0.3797 \\
Margin                      & $+0.0977$ (6.0$\sigma$) & $+0.1410$ (7.2$\sigma$)
                            & $+0.2301$ (25.5$\sigma$) \\
Macro, 7 informative        & 0.4166 & 0.4872 & 0.5560 \\
\textbf{Classes won, every seed} & \textbf{2 of 7} & \textbf{3 of 7} & \textbf{5 of 7} \\
Not beating control & \begin{tabular}[t]{@{}l@{}}Infiltration,\\Nodule,
                        Pneumonia\end{tabular}
                    & \begin{tabular}[t]{@{}l@{}}Infiltration,\\Pneumonia\end{tabular}
                    & none \\
\midrule
\multicolumn{4}{l}{\emph{Intersection over union, $T = 0.1$, the protected metric}} \\
Macro, all 8 classes        & 0.4844 & 0.5325 & 0.5978 \\
Strongest control           & 0.3714 & 0.3631 & 0.3247 \\
Margin                      & $+0.1130$ (5.9$\sigma$) & $+0.1694$ (7.7$\sigma$)
                            & $+0.2731$ (41.4$\sigma$) \\
\textbf{Classes won, every seed} & \textbf{2 of 7} & \textbf{4 of 7} & \textbf{6 of 7} \\
Not beating control & Nodule, Pneumonia & none & none \\
\botrule
\end{tabular}
\end{table*}

Five results show what the aggregate score does and does not measure.

\textbf{The macro overstates grounding.} Cardiomegaly's control reaches 1.0000,
so the class is solved by a fixed box and contributes nothing. Removing it drops
the class activation macro from 0.4896 to 0.4166. At $T(\mathrm{IoBB}) = 0.1$
three classes under that readout, and two under the gradient-weighted one, do not
beat their own control at all; on the protected metric the counts are two and
zero. Which classes fail therefore depends on the overlap measure, which is
itself a reason to report both.

\textbf{The macro is concentrated.} On DenseNet-121, Effusion alone carries 46
percent of the class activation margin and 33 percent of the gradient-weighted
one. We measured this on that classifier and do not claim it for the other two.

\textbf{The count of classes won is the primary summary.} We call the remaining
seven informative, meaning that a fixed box does not already satisfy the metric
on them. After a Holm correction across those seven, at $T(\mathrm{IoBB}) = 0.1$
class activation mapping wins 2 of them in every seed and gradient-weighted
mapping 3; on the protected metric the counts are 2 and 4. Classes significant in
only one seed of three are reported as not established. The one class the
correction changes is Nodule under gradient-weighted mapping at
$T(\mathrm{IoBB}) = 0.1$, where the uncorrected count reads 4. It enters the Holm
sequence at rank 4 of 7 in every seed, with raw $p = 0.0020 / 0.0078 / 0.0156$,
so the corrected values are $0.0078 / 0.0331 / 0.0625$ and the third seed does not
clear $\alpha = 0.05$. On the protected metric the same class survives in all
three seeds, which is why the counts differ by one between the two halves of
Table~\ref{tab:prior}. The per-class test is conservative
rather than exact, because the strongest control is selected on the same sample it
is then tested against; simulating the whole procedure under the null gives a
type-I error of 0.0010 to 0.0012 against 0.019 to 0.021 for a pre-specified
comparator, so these counts are floors.

\textbf{Absolute accuracy and positional signal are different things.} Nodule
under gradient-weighted mapping sits at 0.1308, yet 7 to 10 discordant boxes
favour the model across the three seeds and \emph{none} favours the control in
any of them. A class can carry a real positional signal at a near-floor score.

\textbf{The same diagnostic, applied to PACE.} A measurement of how much of a
localization score is a position prior is worth little if it is run only on the
methods a paper competes with, so the last column of Table~\ref{tab:prior} runs
it on PACE. Against its own size-matched controls PACE wins 5 of the 7 informative
classes in every seed at $T(\mathrm{IoBB}) = 0.1$ and 6 of 7 on the protected
metric, against 2 and 3, and 2 and 4, for the two baselines, and it is the only
readout of the three that beats its own control on every informative class. The
margins are not comparable across the columns and the counts are, for the reason
given in Section~\ref{subsec:metrics}: a control is size-matched to the readout it
tests, so a smaller box weakens the null and inflates the margin, while the count
of classes won is a paired test inside one readout and is untouched by box area.
\emph{That last count is the reading under which the rest of the paper should be taken.} The
diagnostic says that a macro on this benchmark is largely a position prior and
that the standard readouts clear it on two to four findings of seven,
depending on the readout and the overlap measure; the gain
reported below is worth reporting because it moves that count, not because it
moves a macro that a fixed box already partly satisfies.

\begin{table*}[!tp]
\caption{The position-null diagnostic on all three classifiers trained here, at
$T = 0.1$, three seeds. Each cell is the number of the seven informative findings
on which that readout beats its own size-matched control in every seed after a
Holm correction, with the macro margin beside it. IoBB is intersection over the
predicted box and IoU intersection over union. Every column is a within-readout
paired test, so the box-size confound of Section~\ref{subsec:metrics} cannot reach
these counts and they are comparable across columns where the margins are not. The
ordering is the same on both overlap measures and on all three networks, which is
what makes it a statement about the benchmark rather than about one model. Every
cell uses the largest-connected-component box rule; Section~\ref{subsec:boxrule}
reports what happens under two others.}
\label{tab:priorall}
\small\setlength{\tabcolsep}{4pt}
\begin{tabular}{llccc}
\toprule
Metric & Classifier & CAM & Grad-CAM & PACE \\
\midrule
\multirow{3}{*}{IoBB}
 & DenseNet-121  & 2 of 7 ($+0.098$) & 3 of 7 ($+0.141$) & \textbf{5 of 7} ($+0.230$) \\
 & ResNet-50     & 2 of 7 ($+0.102$) & 2 of 7 ($+0.108$) & \textbf{4 of 7} ($+0.205$) \\
 & ConvNeXt-Tiny & 3 of 7 ($+0.137$) & 4 of 7 ($+0.191$) & \textbf{7 of 7} ($+0.288$) \\
\midrule
\multirow{3}{*}{IoU}
 & DenseNet-121  & 2 of 7 ($+0.113$) & 4 of 7 ($+0.169$) & \textbf{6 of 7} ($+0.273$) \\
 & ResNet-50     & 3 of 7 ($+0.115$) & 3 of 7 ($+0.139$) & \textbf{5 of 7} ($+0.242$) \\
 & ConvNeXt-Tiny & 3 of 7 ($+0.157$) & 4 of 7 ($+0.219$) & \textbf{7 of 7} ($+0.334$) \\
\botrule
\end{tabular}
\end{table*}

\textbf{And it replicates on a second corpus, with a different annotation type.}
Everything above is one annotation set. CheXlocalize
\citep{bib-chexlocalize} supplies a second: radiologist \emph{pixel
segmentations} on CheXpert's official validation and test images, drawn by two
board-certified radiologists, which we convert to boxes by the connected-component
rule declared in Section~\ref{sec:instrument}. The images are disjoint from the
CheXpert training pool our classifiers use, verified at zero patient overlap
against all three splits, and all ten annotated findings are already in the
14-class CheXpert head, so no cross-dataset label mapping is involved.
Table~\ref{tab:chexloc} applies the same diagnostic to 2,744 boxes over 584
images.

\begin{table*}[!tp]
\caption{The position null on CheXlocalize, 2,744 boxes over 584 images, three
classifier seeds, connected-component box rule. Counts are informative findings
beaten in every seed after a Holm correction, as in Table~\ref{tab:prior}. The
denominator differs by readout and metric because a class is dropped as saturated
when its own control reaches the ceiling, so a count is read against its own
denominator and not across them. Margins are NOT comparable across columns: each
control is size-matched to its own readout, and PACE emits the smallest box here
(Section~\ref{subsec:areamatch}).}\label{tab:chexloc}
\begin{tabular}{lccc}
\toprule
& Class activation & Gradient-weighted & PACE \\
\midrule
Macro, $T(\mathrm{IoBB}) = 0.1$      & 0.6526 & 0.5975 & 0.6583 \\
Strongest position null              & 0.6059 & 0.6064 & 0.5549 \\
Margin                               & $+0.0466$ & $-0.0088$ & $+0.1034$ \\
\midrule
\textbf{Classes won, IoBB}           & \textbf{2 of 8} & \textbf{2 of 8} & \textbf{3 of 8} \\
\textbf{Classes won, IoU}            & \textbf{3 of 8} & \textbf{3 of 10} & \textbf{4 of 9} \\
\bottomrule
\end{tabular}
\end{table*}

\textbf{Gradient-weighted mapping's macro sits below its own position null}
here, $-0.0088 \pm 0.0155$: a size-matched box that keeps the readout's own
dimensions and only moves scores higher than the readout does. The two baselines
clear the null on two of eight findings at $T(\mathrm{IoBB}) = 0.1$ and three at
$T(\mathrm{IoU})$, against two to four of seven on ChestX-ray14, so the central
measurement of this paper is not a property of one benchmark's annotation style.
PACE's lead narrows: three and four here against five and six on DenseNet-121,
which we report as measured.

\textbf{The diagnostic replicates and so does the ordering.}
Table~\ref{tab:priorall} runs it on all three classifiers and both overlap
measures. The two baselines clear their own position null on two to four of the
seven informative findings in every case, which is the measurement this section
exists to make and which no chest radiograph study we are aware of reports; PACE
clears it on four to seven, and on ConvNeXt-Tiny on all seven. The count is a
within-readout paired test, so unlike a margin it is comparable down a column, and
it is the statistic this paper puts forward. Two conditions travel with it. Every
cell uses the conventional box rule, and Section~\ref{subsec:boxrule} shows that
PACE's count is the one of the three that depends on it. And the seven informative
findings exclude Cardiomegaly, whose control is saturated, so a perfect score here
is seven and not eight.

\noindent\textbf{Do the labels supply the positional signal?}\quad We then build a readout that never sees a pathology label, by scoring each patch
against the same patch position in a reference population of radiographs through
a frozen in-domain backbone. The operator is stated as Eq.~(\ref{eq:anomaly})
where the method is defined; here only the property matters, that no pathology
label enters it. It reaches a class macro of
0.5036 at $T(\mathrm{IoU}) = 0.1$ against 0.4844 for class activation mapping and
0.5325 for gradient-weighted mapping on DenseNet-121. Its own value does not
depend on that classifier at all, since neither the frozen backbone nor the
reference bank reads the classifier: its classifier-seed spread is exactly zero
(Table~\ref{tab:ablation}), and the classifier is named only because it fixes
what it is being compared against. The label-free readout lands in the same
accuracy range as a label-trained one, but the measurements do not establish
equivalence. Against class activation mapping the three seeds give
$p = 0.066 / 0.599 / 0.325$, and against gradient-weighted mapping
$0.855 / 0.0039 / 0.024$, so the latter significantly wins two seeds of three.
An equivalence claim would need a declared margin and a two one-sided test
\cite{bib-schuirmann1987}; running one at the 0.05 margin this project already
declares as its smallest interesting difference, equivalence is supported in
\textbf{one} of the six testable cells on the protected metric, against class
activation mapping at $T(\mathrm{IoU}) = 0.25$, and in none against
gradient-weighted mapping. Appendix~\ref{app:tost} gives the cell bookkeeping,
including the floor-effect exclusions and the one apparent second cell that the
box-size confound disqualifies. These paired tests reach 80 percent power only at
an absolute difference of about 0.044 to 0.048, so smaller differences are
unresolved rather than absent.

Two observations bound the interpretation. The three label-free readouts differ
from \emph{each other} far more than any of them differs from class activation
mapping: pooled over the 984 boxes at $T(\mathrm{IoU}) = 0.1$ they sit at 0.5396,
0.3496 and 0.1535, the two radiograph-pretrained backbones separating by 0.190 at
worst-seed $p = 1.3 \times 10^{-32}$ and the natural-image one \cite{bib-dinov2}
by 0.386 at $p = 4.8 \times 10^{-96}$. On macros, which are a different quantity
and not a paired difference, the same gaps are 0.180 and 0.362, from 0.5036,
0.3232 and 0.1413. DINOv2 localizes these findings worse than its own box fixed at
the image centre, 0.1413 against 0.3616. That ordering does not identify
pretraining domain as the cause, since the three differ in objective, corpus size
and recipe as well as in domain; what it shows is that the choice of frozen
representation dominates, and that a positional signal comparable to a class
activation map's can be obtained without labels. Neither statement identifies what
CAM's own positional signal is made of, which would need the labels varied with
everything else held fixed. That is the licence for Eq.~(\ref{eq:pace}).

A third measurement licenses combining the streams rather than replacing one with
the other: over the 984 boxes the two are 23.6 percent discordant, and an oracle
that picked the better of the two per box would reach 0.6514. All three of those
figures are pooled over the boxes, since the oracle is a per-box choice, and the
comparator is likewise pooled and taken per seed: in each seed we take whichever
stream is better \emph{in that seed}, which averages to 0.5427 and is neither
stream's own mean, so the headroom is 0.109. The corresponding class macros, which
Table~\ref{tab:ablation} reports, are lower for every quantity here and must not be
subtracted from these. They fail on different images.

\section{The human ceiling}\label{sec:human}

Section~\ref{subsec:rq1} says that the readouts clear their own position null on
a minority of findings. That is only worth reporting if the null is beatable at
all. A diagnostic on which nothing can succeed measures the difficulty of the
task, not the quality of the method, and this benchmark gives no way to tell the
two apart: ChestX-ray14 carries a single annotation per finding and no second
reader.

CheXlocalize does. Its test images carry two independent sets of radiologist
segmentations: a ground truth from two board-certified radiologists, and a
\emph{human benchmark} drawn by a separate group of three, which
\citet{bib-chexlocalize} introduce as the ceiling that saliency methods fall short of.
Nobody has asked what that ceiling scores against a position null.

We ask it with the same construction used for every model row, substituting one
term. The prediction is the human benchmark's own box; the target is the
ground-truth box; the controls are the same three nulls, size-matched to
\emph{the human's} box and differing from it only in position. The result is
therefore on the same axis as Table~\ref{tab:prior}: whatever part of the human
score a displaced copy of the human's own box already achieves is not evidence of
localization by anybody.

\textbf{What this is and is not.} It is the same position-null diagnostic applied
to the strongest annotator available, so that the model rows have a scale. It is
\emph{not} an upper bound on achievable performance, because the target is
another radiologist group's segmentation rather than pathology ground truth, and
assigning one group as prediction and the other as target is a construction of
ours that the two do not stand in symmetrically. Nor is it an inter-reader
agreement statistic, which would involve no position null at all. We therefore
call it a ceiling only in the narrow sense of the highest position-null margin
any annotator in this corpus achieves.

\begin{table*}[!tp]
\caption{The position null applied to human readers, and what it costs to
perturb the two constructions the comparison rests on. Three board-certified
radiologists' segmentations scored against a separate two-reader set, on the
1,294 (image, finding) pairs both marked. \emph{Reversed} swaps which set is the
prediction and which the target, a choice that is ours and not a property of the
data. \emph{Union} takes the bounding box of the whole mask instead of one box
per connected component, on both sides and on the model rows alike. Counts are
informative findings beaten in every control draw after a Holm correction; the
denominator changes with the rule because saturation is judged on the
control.}\label{tab:human}
\begin{tabular}{llcc}
\toprule
& & $T(\mathrm{IoBB}) = 0.1$ & $T(\mathrm{IoU}) = 0.1$ \\
\midrule
\multirow{3}{*}{component rule}
  & Class activation mapping  & 2 of 8 & 3 of 8 \\
  & Gradient-weighted mapping & 2 of 8 & 3 of 10 \\
  & PACE                      & 3 of 8 & 4 of 9 \\
\cmidrule(l){2-4}
  & \textbf{Human, as reported} & \textbf{4 of 8} & \textbf{7 of 8} \\
  & Human, reversed            & 4 of 8 & 4 of 8 \\
\midrule
\multirow{3}{*}{union rule}
  & Class activation mapping  & 2 of 7 & \\
  & Gradient-weighted mapping & 2 of 7 & \\
  & PACE                      & 3 of 7 & \\
\cmidrule(l){2-4}
  & \textbf{Human}             & \textbf{5 of 7} & \textbf{5 of 7} \\
  & Human, reversed            & 4 of 7 & 4 of 7 \\
\bottomrule
\end{tabular}
\end{table*}

\textbf{Radiologists clear the null and the readouts do not, in all four
constructions.} The human beats its own position null on four to seven of the
seven or eight informative findings depending on the box rule and on which
annotator set is treated as the prediction, where the readouts reach two to four.
\emph{The single most favourable cell is seven of eight}, on the protected metric
under the component rule in the reported direction, with a macro margin of
$+0.2665$ to $+0.2857$ and a bootstrap interval of $[+0.234, +0.325]$; reversing
the roles takes it to four of eight and the union rule to five of seven. We lead
with the range because a cross-validation asked for these two perturbations and
the headline did not survive either of them intact. So on these pairs the null is beatable by a wide margin, and a
readout that fails to beat it is telling us about the readout rather than about
an impossible task. The conditioning is not incidental and is quantified two
paragraphs below. This is the sentence that Section~\ref{subsec:rq1} cannot
supply on its own, and it is available on no other chest-radiograph corpus we
know of.

\textbf{The forward and reversed numbers differ for a reason this paper has
already measured, which is why we can report the asymmetry rather than merely
disclose it.} The two annotator sets are not exchangeable. \citet{bib-chexlocalize}
record that the ground truth was drawn by two radiologists of 18 and 27 years'
experience and the benchmark by three certified in Vietnam with 9, 10 and 18
years, given the same contouring instructions; neither passage describes any
adjudication of one reader's contour against another's, so there is no consensus
step for us to model. What we can measure is the consequence. On the 1,294
shared pairs the benchmark's mask is the smaller one, at a median $0.682$ of the
ground truth's area and larger on only 14.7 percent of pairs; in absolute terms
the medians are 0.115 and 0.194 of the image.

That is exactly the mechanism of Section~\ref{subsec:areamatch}, operating on
human annotations. Scoring the smaller set against the larger, our reported
direction, pairs a tight box with a generous target and a position null that is
size-matched to the tight box and therefore weak. Reversing it pairs a generous
box with a tight target and a null that, being generous too, still overlaps
often. \emph{The 7 of 8 and the 4 of 8 are the same annotations read at two box
sizes}, and neither is more correct than the other, which is why
Table~\ref{tab:human} reports both and the surrounding text quotes the range.
This is also, we think, the strongest single illustration in the paper that the
box-area effect is not an artefact of machine readouts: it is visible between two
groups of radiologists given identical instructions.

\textbf{The metric matters here too, and in the direction the rest of the paper
predicts.} On intersection over the predicted box the human clears only four of
eight, against seven on intersection over union. A human draws a tight
segmentation, so the statistic that divides by the predicted area alone
penalises exactly the behaviour that makes the annotation good. That is the same
mechanism as Section~\ref{subsec:areamatch}, seen on a reader rather than on a
model.

\textbf{Two limits, both of which make the ceiling easier than the model rows and
so are stated with it.} The pairs are those \emph{both} annotator sets marked,
1,294 of 1,331, discarding 37 that only the ground truth marked and 3 that only
the human benchmark did; selecting on agreement removes the cases the two readers
disagreed about, and the model rows are not selected that way. And where a human
drew several components for one finding we score their largest against each
ground-truth component separately, which is exactly what every model readout here
undergoes: the evaluation also emits one box per (image, finding) and scores it
against each annotated component. \emph{That second choice does not favour the
human}, and an earlier draft of this section said it did. It penalises
multi-instance findings, for the human and the models alike, since one box cannot
cover a bilateral finding's two components; what it buys is that both are held to
the same convention. So of the two qualifications, the first makes the human's
task easier and the second makes it no easier than the models'.

\section{Protocol dependence: three choices that move a number more than a method does}\label{sec:protocol_dep}

\subsection{Predicted box area}\label{subsec:areamatch}

The readouts compared in Table~\ref{tab:main} emit boxes of markedly different
sizes at the fixed extraction threshold $\tau = 0.5$, and a product of two maps
normalised to $[0,1]$ concentrates by construction, so a reviewer is entitled to
ask whether PACE finds the finding more often or merely draws a smaller box. We
answer by sweeping $\tau$ for each readout, which changes the box area without
touching the map, and comparing the three at matched area.

\begin{table*}[!tp]
\caption{Box-extraction threshold against box area and accuracy, DenseNet-121,
three seeds, pooled over the 984 boxes. $\tau$ is the fraction of each map's own
range above which a pixel joins the mask, and the area column is the median
predicted-to-annotated area ratio recovered from the two overlap measures. The
convention $\tau = 0.5$ is \textbf{not} the best setting for either baseline, and
that is why a comparison at fixed $\tau$ is not a comparison at matched area.
Rows in bold are the ones matched to PACE's area and used in
Table~\ref{tab:areamatch}.}
\label{tab:sweep}
\begin{tabular}{lccccc}
\toprule
Readout & $\tau$ & Area & $T = 0.1$ & $T = 0.25$ & $T = 0.5$ \\
\midrule
\multirow{3}{*}{CAM}
 & 0.50 & 5.92$\times$ & 0.5274 & 0.3049 & \textbf{0.1406} \\
 & 0.70 & 2.09$\times$ & \textbf{0.5884} & 0.3726 & 0.0705 \\
 & \textbf{0.75} & \textbf{1.45$\times$} & 0.5857 & 0.3591 & 0.0593 \\
\midrule
\multirow{3}{*}{Grad-CAM}
 & 0.50 & 2.80$\times$ & 0.5691 & 0.3526 & \textbf{0.1230} \\
 & 0.60 & 1.78$\times$ & \textbf{0.5813} & 0.3662 & 0.0725 \\
 & \textbf{0.65} & \textbf{1.36$\times$} & 0.5783 & 0.3543 & 0.0532 \\
\midrule
\textbf{PACE} & \textbf{0.50} & \textbf{1.40$\times$} & \textbf{0.6358} & \textbf{0.4096} & 0.0972 \\
\botrule
\end{tabular}
\end{table*}

\begin{table*}[!tp]
\caption{PACE against each baseline at \emph{matched box area}, and against each
baseline at that baseline's own best threshold, DenseNet-121, three seeds,
intersection over union, pooled paired differences over the 984 boxes. The upper
number of each pair is the mean over three classifier seeds. The bracket below it
is a \emph{different quantity}, on the same convention as Table~\ref{tab:main}:
the 95 percent image-clustered interval in the governing seed, which is the seed
whose evidence against zero is weakest. It is not an interval for the mean above
it and need not contain it. Every entry is positive and every interval excludes
zero, including at $T = 0.5$, where the fixed-threshold comparison of
Table~\ref{tab:main} reverses. The ``its best $\tau$'' rows compare against each
baseline at the threshold that maximises that baseline's own accuracy at
$T = 0.1$, and carry no interval because the point of them is only that the two
framings agree.}
\label{tab:areamatch}
\small\setlength{\tabcolsep}{3pt}
\begin{tabular}{llccc}
\toprule
Comparison & Baseline at & $T = 0.1$ & $T = 0.25$ & $T = 0.5$ \\
\midrule
\multirow{3}{*}{PACE over CAM}
 & matched area & $+0.050$ & $+0.051$ & $+0.038$ \\
 &              & \scriptsize $[+.017,+.060]$ & \scriptsize $[+.016,+.067]$ & \scriptsize $[+.010,+.049]$ \\
 & its best $\tau$ & $+0.047$ & $+0.037$ & $+0.027$ \\
\midrule
\multirow{3}{*}{PACE over Grad-CAM}
 & matched area & $+0.058$ & $+0.055$ & $+0.044$ \\
 &              & \scriptsize $[+.010,+.053]$ & \scriptsize $[+.015,+.064]$ & \scriptsize $[+.014,+.052]$ \\
 & its best $\tau$ & $+0.055$ & $+0.043$ & $+0.025$ \\
\botrule
\end{tabular}
\end{table*}

\textbf{The gain is not box size, and part of it was.} Against class activation
mapping the difference at $T(\mathrm{IoU}) = 0.1$ falls from $+0.108$ at fixed
$\tau$ to $+0.050$ at matched area, so roughly half of the fixed-threshold margin
was the baseline's oversized box; against gradient-weighted mapping it falls only
from $+0.067$ to $+0.058$. Both survive in every seed and under the clustered
interval, and comparing each readout at its own best threshold instead gives
$+0.047$ and $+0.055$, so the two framings agree.

\textbf{At strict overlap the ordering depends on the operating point, and we
state that rather than resolve it.} Table~\ref{tab:main} reports PACE losing to
class activation mapping at $T(\mathrm{IoU}) = 0.5$. At matched area the sign is
the other way, $+0.038$ and $+0.044$ with intervals excluding zero. Both are true
of what they measure and they are not the same comparison. What
Table~\ref{tab:sweep} shows is that the baselines reach their best strict-overlap
accuracy at $\tau = 0.5$, where their boxes are 4.2 and 2.0 times PACE's, and
that shrinking them to PACE's area costs 0.081 and 0.070 there; PACE reaches its
own best at 0.5 as well. \emph{So the honest statement is that PACE has the higher
strict-overlap accuracy at equal box area and the lower one at each readout's
default threshold, and this paper does not establish which comparison should
govern.} Deciding it would need the extraction threshold chosen for the strict
metric on data other than these 984 boxes, and no such data exists on this
benchmark. What does not survive is the earlier unconditional reading, that PACE
finds the right region without fitting it better; that was stated as a property
of the method and it is a property of the threshold. The matched-area advantage
does not extend to $T(\mathrm{IoU}) = 0.75$, where both intervals contain zero
and every readout is at the floor.

\textbf{The cost is a sensitivity we have to declare.} PACE's accuracy varies far
more with $\tau$ than either baseline's: over the sweep it spans 0.083 to 0.636
where class activation mapping spans 0.352 to 0.588. The mechanism is not
monotone shrinkage. Between $\tau = 0.50$ and $\tau = 0.55$, 58 percent of
predicted boxes \emph{grow}, by a median factor of 1.85 and by up to 14.9. The
denominator is worth stating: a predicted box's area is not recorded and is
recovered from the two overlap measures, which is possible only where both are
positive, so the fraction is over the 1{,}640 of 2{,}952 box instances where the
area is recoverable at both thresholds, and the boxes that miss the annotation
entirely are not among them. A box that grows when the threshold rises cannot be
explained by the mask shrinking, and the natural reading is that the product map
is peaked and multi-modal and the largest connected component moves to another
lobe; we have not opened the maps to confirm it, so it is a reading and not a
measurement. Either way the convention $\tau = 0.5$ sits in a stable region for
PACE by inheritance rather than by design, and a readout whose box identity is
discontinuous in its own extraction threshold is a weaker object than one whose is
not. A box generator less brittle than largest-connected-component is the obvious
repair, and Section~\ref{subsec:boxrule} reports two.

\textbf{On the second corpus the confound runs the other way, which is the
sharpest evidence we have that it is a confound and not a property of the
method.} Recovering predicted areas the same way on CheXlocalize gives class
activation mapping $3.56\times$ the target, its gradient-weighted variant
$1.40\times$, and PACE $0.76\times$: PACE \emph{undershoots} here where it
overshot on ChestX-ray14. At $T(\mathrm{IoU}) = 0.1$ a box that merely contains
the target scores $1/3.56 = 0.28$ with no alignment at all, so the loose
threshold now rewards the larger box, and at the shared extraction threshold PACE
\emph{loses} to class activation mapping by $-0.056$ with an image-clustered
interval excluding zero. Recalibrating each readout to emit $0.70\times$ to
$0.88\times$ reverses that to $+0.013$, not significant, and moves the gain to
the strict thresholds, $+0.075$ at $T(\mathrm{IoU}) = 0.25$ and $+0.028$ at
$0.5$. \textbf{The same comparison, on the same boxes, reads $-0.056$ or
$+0.013$ depending only on a threshold neither number reports.} A localization
comparison comes with a box-area ratio or it is partly a claim about box size.

\subsection{The mask-to-box rule}\label{subsec:boxrule}

Section~\ref{subsec:areamatch} found PACE's box identity discontinuous in the
extraction threshold and named the largest-connected-component rule as the
suspect. That rule is a convention, not a measurement, so we replaced it and
report what happened. Two alternatives were declared on stated grounds before
either was run.

\emph{Union}: the bounding box of \emph{every} pixel above the threshold. The
selected set shrinks monotonically as the threshold rises, so this box can only
shrink and the discontinuity cannot occur. \emph{Peak}: the connected component
containing the map's global maximum. The argmax pixel is selected at every
threshold below one, so the component's identity moves continuously while
single-lobe selection is kept, which \emph{union} gives up. We record the order
honestly: \emph{peak} was added after \emph{union} was measured and seen to
convert multi-modality into a large box. All three are reported.

\begin{table*}[!tp]
\caption{The same three readouts under three mask-to-box rules, DenseNet-121,
three seeds, $\tau = 0.5$, pooled over the 984 boxes. \emph{Largest} is the
convention and is what every other number in this paper uses. The last column is
the spread of $T = 0.1$ accuracy across the three rules, which is the quantity
this table exists to show.}
\label{tab:boxrule}
\small\setlength{\tabcolsep}{3pt}
\begin{tabular}{llcccc}
\toprule
Readout & Rule & Area & $T = 0.1$ & $T = 0.25$ & $T = 0.5$ \\
\midrule
\multirow{3}{*}{CAM}
 & largest & 5.92$\times$ & 0.5274 & 0.3049 & 0.1406 \\
 & peak    & 5.94$\times$ & 0.5217 & 0.3001 & 0.1382 \\
 & union   & 7.77$\times$ & 0.5068 & 0.2757 & 0.1369 \\
\midrule
\multirow{3}{*}{Grad-CAM}
 & largest & 2.80$\times$ & 0.5691 & 0.3526 & 0.1230 \\
 & peak    & 2.79$\times$ & 0.5606 & 0.3472 & 0.1220 \\
 & union   & 4.03$\times$ & 0.5569 & 0.3303 & 0.1216 \\
\midrule
\multirow{3}{*}{\textbf{PACE}}
 & largest & 1.40$\times$ & \textbf{0.6358} & \textbf{0.4096} & 0.0972 \\
 & peak    & 1.98$\times$ & 0.4201 & 0.2243 & 0.0512 \\
 & union   & 6.64$\times$ & 0.4404 & 0.2029 & 0.0451 \\
\midrule
\multicolumn{3}{l}{\emph{spread across rules at $T = 0.1$}}
 & CAM 0.021 & Grad-CAM 0.012 & \textbf{PACE 0.216} \\
\botrule
\end{tabular}
\end{table*}

\textbf{The result is negative and we report it as the principal limitation of
this paper.} The two baselines are almost invariant to the rule, moving 0.021 and
0.012 at $T(\mathrm{IoU}) = 0.1$. PACE moves 0.216, and under either alternative
it loses to both baselines by margins that are significant in every seed and under
the image-clustered interval: against class activation mapping $-0.096$ under
\emph{peak} and $-0.051$ under \emph{union} at $T = 0.1$, and against the
gradient-weighted variant $-0.132$ and $-0.101$. \emph{So the gain reported in
Table~\ref{tab:main} and Table~\ref{tab:areamatch} is a property of the product map
together with the largest-component rule, and not of the product map alone.} A
reader who adopts a different box generator should not expect it.

\textbf{Two things the same table measures, which are worth separating from the
verdict.} First, it supplies the evidence that Section~\ref{subsec:areamatch}
lacked for calling the product map multi-modal: the ratio of the union box to the
largest-component box at the same threshold is 1.31 for class activation mapping
and 1.44 for the gradient-weighted variant, against \textbf{4.74} for PACE. A map
whose all-component box is nearly five times its largest-component box has its
mass in several separated lobes, and that is read off committed artifacts rather
than asserted. Second, \emph{peak} is the more informative of the two failures.
It keeps single-lobe selection and still costs PACE 0.216, which says that the
component holding the product's global maximum is often not the component that
contains the finding. The reading we take from this, and it is a reading rather
than a measurement, is that the product's argmax is sharp and unreliable while
its broadest region of agreement is not, so the largest-component rule is doing
substantive work for this readout rather than merely extracting a box. If that is
right, the honest specification of the method includes the rule, and the
successor worth building is a readout whose peak can be trusted.

\begin{figure*}[!tp]
\centering
\includegraphics[width=\textwidth]{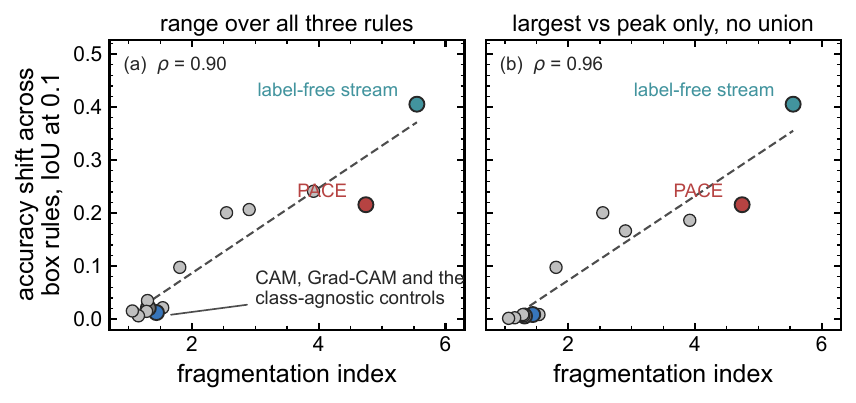}
\caption{Sensitivity to the mask-to-box rule for thirteen DenseNet-121 readouts
at three seeds. The fragmentation index is the median all-component box area
divided by the median largest-component box area; it increases when thresholded
regions lie farther apart. The vertical axis is the range of pooled accuracy at
$T(\mathrm{IoU}) = 0.1$ across the three rules in Table~\ref{tab:boxrule}.
Panel (a) partly shares construction between its axes. Panel (b) instead compares
the index with the largest-to-peak difference, which uses no union geometry and
gives $\rho = 0.96$. Dashed lines are least-squares fits over the measured range.
Fragmentation does not imply poor localization; it predicts dependence on the
box rule.}
\label{fig:boxrule}
\end{figure*}

\textbf{The sensitivity is not arbitrary: it tracks how fragmented the thresholded
mask is.} Running the three rules over all thirteen readouts in this paper gives
Figure~\ref{fig:boxrule}. The shift a readout suffers when the box rule changes
rises with its fragmentation index at Spearman $\rho = 0.90$, and at
$\rho = 0.96$ under the control that removes the shared construction between the
two axes. The two ends are interpretable. Class-blind class activation mapping is
at 1.06, a single blob, and moves 0.015; the label-free stream on its own is at
5.55 and moves 0.406, the largest shift we measure, with PACE between them at 4.74
and 0.216. Every rule that combines two maps sits high, because thresholding a
product or a mean leaves pieces wherever either factor is strong.

\emph{The observation is not about PACE, and we are careful about how far it
generalises.} Thirteen readouts on one architecture, most of them derived from the
same class activation map or the same frozen backbone, are not thirteen
independent samples, and this is a diagnostic relationship rather than a law. What
it does support is a practical claim we would make on one point alone: a
localization number here that does not state its mask-to-box rule cannot be
compared with one that does, and the cost of ignoring the rule is 0.012 to 0.021
for the diffuse maps of 2016 and 2017 and 0.216 for a concentrated one. We found
it because our own readout is concentrated; a paper reporting only class
activation mapping would have measured 0.021 and concluded the rule is
immaterial.

\section{Overlap is not faithfulness}\label{subsec:analysis}

\noindent\textbf{Does the map describe the classifier?}\label{subsec:faithful}\quad Overlap with a radiologist box cannot detect a map that lands on plausible
anatomy for reasons unrelated to the prediction, which is the criticism of
\cite{bib-grounding}. We test this with the occlusion measurement introduced
by \cite{bib-zeiler2014}: occlude each cell of a $16 \times 16$ grid with the
image mean, record the change in the class score, and correlate that causal map
against the readout by Spearman rank over the same 984 boxes. Table~\ref{tab:faithful} reports three seeds on all three
architectures.

\begin{table*}[!tp]
\caption{Occlusion faithfulness, Spearman rank correlation between each readout
and the classifier's own causal map, three seeds, 984 boxes. Higher is more
faithful. The lower block is quoted from \cite{bib-grounding}, measured on other
models and another corpus, and includes that paper's own positive controls, which
are a DenseNet-121 and a ResNet-50 read out with gradient-weighted mapping. Our
values for the same readout on those two architectures, 0.2047 and 0.1424,
reproduce theirs rather than extend them; the PACE column is the new one.}
\label{tab:faithful}
\begin{tabular}{lccc}
\toprule
Classifier & CAM & Grad-CAM & PACE \\
\midrule
DenseNet-121   & \textbf{0.2230 (114)} & 0.2047 (152) & 0.1541 (171) \\
ResNet-50      & \textbf{0.1603 (184)} & 0.1424 (167) & 0.1048 (93) \\
ConvNeXt-Tiny  & \textbf{0.1924 (114)} & 0.1771 (71)  & 0.1452 (220) \\
\midrule
\multicolumn{4}{l}{\emph{Reported by \cite{bib-grounding} on other models and another corpus}} \\
Vision-language attention, six probes & \multicolumn{3}{c}{$-0.050$ to $-0.168$} \\
Vision-language attention, LLaVA-RAD & \multicolumn{3}{c}{$+0.137$} \\
Its own CNN controls, Grad-CAM & \multicolumn{3}{c}{$+0.170$ (ResNet-50), $+0.209$ (DenseNet-121)} \\
\botrule
\end{tabular}
\end{table*}

Two things follow, and the first is not ours. The anti-correlation
\cite{bib-grounding} reports is a property of vision-language attention and not of
this family of readouts, and \emph{that paper already shows it}: a DenseNet-121
and a ResNet-50 read out with gradient-weighted mapping are its own positive
controls, at $\rho = 0.209$ and $0.170$. Our values on the same readout, 0.2047
and 0.1424, reproduce those on a different corpus and a different box set with
our own occlusion implementation, which is worth having as an independent check
but is not a new finding and we do not present it as one. Its negative values
likewise belong to a named subset rather than to every model it tests: the five
MedGemma variants and the Qwen3-VL probe run from $-0.050$ to $-0.168$, while
LLaVA-RAD is $+0.137$, which that paper attributes to near-zero causal mass
rather than to grounding. What is new here is the third column. The same
measurement limits the claim for PACE. PACE is the least faithful of the three on all three
architectures and in every seed: class activation mapping beats it by 0.047 to
0.069 at worst-seed $p$ between $3.2 \times 10^{-7}$ and $2.6 \times 10^{-34}$ by
a paired signed-rank test \cite{bib-wilcoxon1945}.
This follows from the construction rather than contradicting it, since PACE
multiplies in a stream the classifier has never seen, so the product moves toward
image anomaly and away from classifier evidence. A second statistic, the area
under the deletion curve \cite{bib-rise}, does not order the three readouts stably
across architectures: PACE is last on DenseNet-121 and ResNet-50, on ConvNeXt-Tiny
it beats gradient-weighted mapping at worst-seed $p = 0.0015$ and is not separated
from class activation mapping, and the two baselines swap places between the first
two architectures. We report that instability rather than quote the statistic where
it agrees with us.

PACE's map is therefore a localizer, not an explanation of the model. It agrees
with the radiologist more at loose overlap thresholds, where its gain
is significant on every classifier, and no better at strict ones; it agrees with
the classifier less everywhere, on all three architectures and in every seed. A
reader who wants to know why the model predicted a finding should read the class
activation map. A reader who wants the region that contains the finding, and can
work at $T(\mathrm{IoU}) = 0.1$ to $0.25$, should read PACE.

\noindent\textbf{Cross-dataset transfer.}\label{subsec:cross}\quad Each classifier is scored on the other corpus without retraining, through the
seven label pairs that are exact or renamed; three CheXpert labels that subsume
ChestX-ray14 findings rather than renaming them are excluded and recorded, and an
absent label is masked rather than scored as a negative. Multi-source pretraining
work resolves synonyms and unions the label sets in the same spirit but encodes no
subsumption \cite{bib-harmoe}, which is precisely the relation that makes three of
our eight boxed findings unusable across corpora. The comparator uses the same
seven classes scored in domain, never a 13- or 14-class mean.
Both directions use DenseNet-121 at three seeds, so the asymmetry is not an
architecture effect. ChestX-ray14 to CheXpert loses 0.109 ($0.7072 \pm 0.0028$
against $0.8163 \pm 0.0006$); CheXpert to ChestX-ray14 loses 0.011
($0.7837 \pm 0.0007$ against $0.7946 \pm 0.0003$). The asymmetry is a factor of
ten at seed spreads of 0.0003 to 0.0028. Candidate causes are the larger CheXpert
training pool, the different mining tools and the differing prevalence of the
shared findings; none is established here, so we report the measurement.

\noindent\textbf{The long tail runs in opposite directions on the two datasets.}\label{subsec:tail}\quad Classes are sorted by training-set positive count and cut at the rank fractions
0.34 and 0.67, which is the same cut the training-time frequency diagnostic uses,
so the two are readable side by side. On ChestX-ray14 that gives a head of
Infiltration (12,375 positives), Effusion (7,682), Atelectasis (7,436) and Nodule
(4,214); a middle of Mass (3,634), Consolidation (2,559), Pneumothorax (2,397),
Pleural Thickening (2,001) and Cardiomegaly (1,542); and a tail of Emphysema
(1,279), Edema (1,242), Fibrosis (1,139), Pneumonia (777) and Hernia (128).
The bands score
head 0.7307 to 0.7592, middle 0.7726 to 0.8110 and tail 0.8070 to 0.8346 across
the three architectures: area under the curve rises monotonically with rarity. CheXpert on the same code runs the ordinary way round, 0.8370 / 0.7982 /
0.7770. The per-class results suggest why: ChestX-ray14's head band carries
Infiltration at 0.68 to 0.70, its hardest and
most frequent finding, while its tail holds visually distinctive findings.
The inversion matters because the long-tailed chest radiograph literature rests on
the opposite premise \cite{bib-ltcxnet,bib-cxrlt2026}: ChestX-ray14 does not support
the claim that rare findings are harder, so a method motivated by that claim
cannot be argued for on this dataset.

\begin{figure*}[!tp]
\centering
\includegraphics[width=0.80\textwidth]{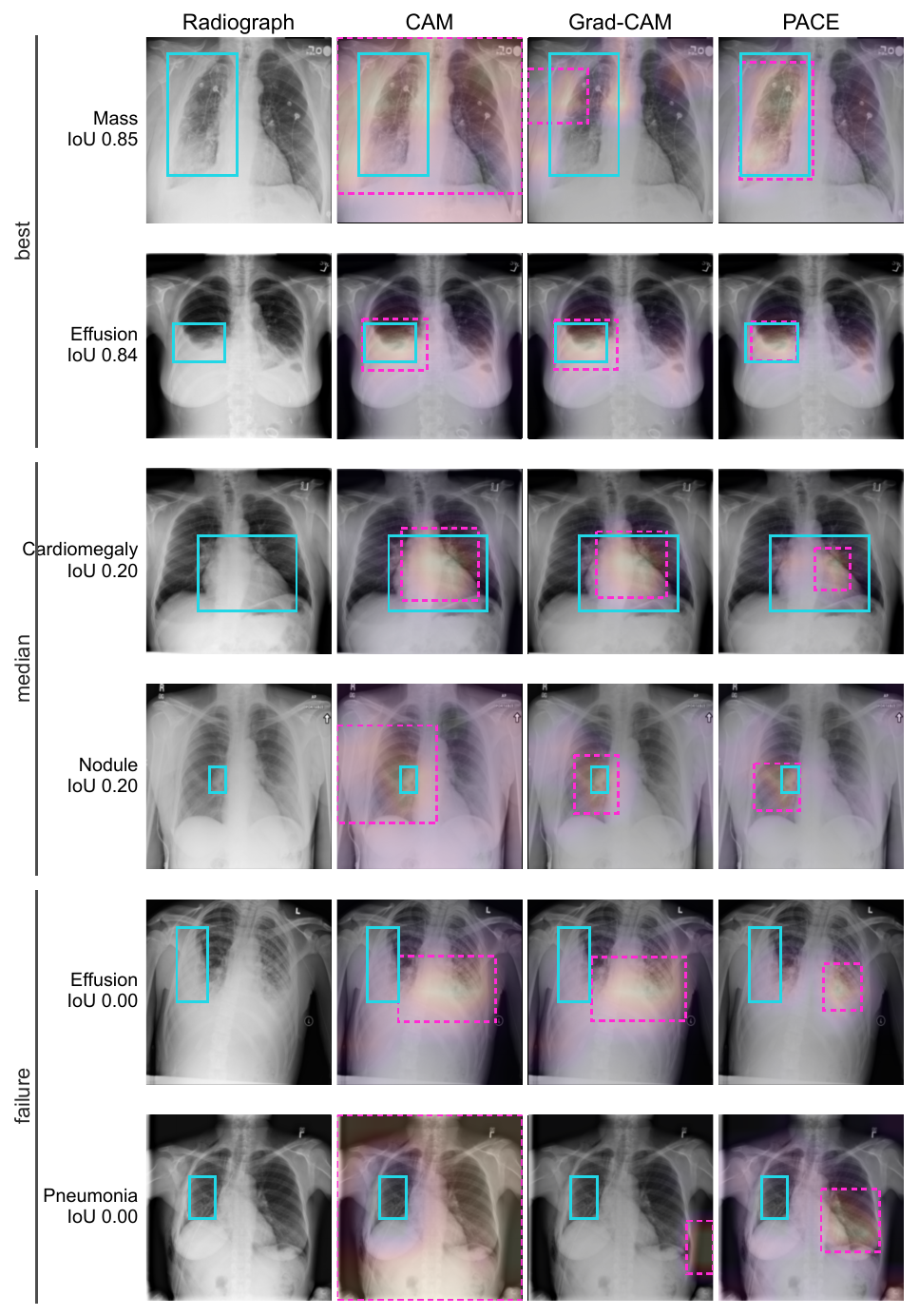}
\caption{Qualitative localization, with failure cases. Rows are one annotated
image and class each; columns are the radiograph and the three readouts on
DenseNet-121. Cyan marks the annotated box, magenta the predicted one, and the
heat map is each readout's own map normalised to its own range and thresholded at
0.5 of that range to form the box. Examples are not selected by eye: they are the
highest, middle and lowest ranks of the measured per-box intersection over union
in the committed per-box record, ranked on the designated bank so that the
selection and the maps come from one evaluation. The failure block is Effusion and
Pneumonia at 0.00, where all three readouts miss the annotated box entirely. The panel also shows the box-size difference discussed in
Section~\ref{subsec:metrics}: the predicted box of PACE is smaller than the class
activation map's on every row of the panel, by factors of 1.9 to 9.4 in area, and
the ordering is measured from the committed box records rather than judged by
eye.}
\label{fig:qual}
\end{figure*}

\noindent\textbf{Qualitative examples.}\label{subsec:qual}\quad Figure~\ref{fig:qual} shows examples selected by rank in the measured per-box
overlap rather than by eye, so the panel is reproducible from a committed record,
and one third of it is failure cases.

\textbf{Both halves replicate on CheXlocalize, including the half that costs us.}
Repeating the occlusion test on the second corpus with the CheXpert classifiers
gives Spearman correlations of 0.2380 for class activation mapping, 0.1970 for
its gradient-weighted variant and 0.1391 for PACE, positive on 90.8, 86.0 and
77.1 percent of boxes, against 0.2230, 0.2047 and 0.1541 on ChestX-ray14. The
ordering is identical and every value is positive, so the criticism that visual
explanations in medical models are anti-correlated with the model's own causal
map does not reach class-activation-family readouts on either corpus. And PACE
remains the least faithful of the three on both, which follows from multiplying
in a stream the classifier never saw: \emph{it is a localizer, not an explanation
of the model}, and the gain in Section~\ref{sec:results} should not be quoted
without this cost beside it.

\section{PACE: the readout these measurements produced}\label{sec:method}

The measurements above say that a localization macro is substantially a position
prior and that the published fusion methods take their second spatial map from the
same image-level labels that trained the first. This section builds the readout
that follows from those two observations. \emph{It is reported here as the case
study that made the protocol dependence of Section~\ref{sec:protocol_dep}
visible}, not as a method established independently of it, and
Section~\ref{sec:confirm} states what its selection was and was not worth.

Let $x \in \mathbb{R}^{H \times W}$ be a radiograph and $y \in \{0,1\}^{C}$ its
image-level label vector over $C$ findings. Training uses pairs $(x, y)$ only. A
network $g_\theta$ produces per-class probabilities
$\hat{p} = \sigma(g_\theta(x)) \in [0,1]^{C}$, trained with a per-class binary
cross entropy

\begin{equation}
\mathcal{L} = -\sum_{c=1}^{C} w_c \big[ y_c \log \hat{p}_c
              + (1-y_c) \log (1-\hat{p}_c) \big],
\label{eq:bce}
\end{equation}

where $w_c$ balances positives against negatives within the batch. At test time
the same network must also emit a per-class map $M_c \in \mathbb{R}^{h \times w}$
and, from it, a box. No box is ever seen in training; the 984 released boxes are
used for evaluation only and lie entirely inside the untouched test split.

\noindent\textbf{Class-evidence stream.}\quad Let $A^k$ be the $k$-th channel of the final feature map and $w^c_k$ the
classifier weight joining channel $k$ to class $c$, so that the class score is
$s_c = \sum_k w^c_k \,\mathrm{GAP}(A^k)$. The class-evidence map is the
projection of those weights back onto the spatial grid,

\begin{equation}
M_c \;=\; \sum_k w^c_k A^k ,
\label{eq:cam}
\end{equation}

which is exact for any architecture that pools and then applies one linear layer
\cite{bib-zhou2016}. Every classifier trained here meets that condition by
construction rather than by luck: the backbone's own classification head, which
for a modern transformer-style network includes a layer normalisation, is
discarded at build time and replaced by global average pooling followed by a
single linear layer, so Eq.~(\ref{eq:cam}) is exact for the convolutional and the
transformer backbones alike and no approximation enters. The one classifier we do
not train is the released checkpoint of Section~\ref{subsec:main}, which keeps its
own head; the exact decomposition for it is derived and checked numerically
there. This stream carries the answer to \emph{which finding}, and
it is the stream that inherits the label process.

\noindent\textbf{Population-anomaly stream.}\quad Let $f$ be a backbone frozen after pretraining on radiographs with no pathology
labels, and let $f_i(x) \in \mathbb{R}^{d}$ be its token at spatial position $i$.
Draw a reference set $\mathcal{R}$ of $N$ unlabelled radiographs. Which split it
comes from is a decision rather than a detail. We draw it from the
\textbf{training pool}, which is disjoint from the box-annotated subset by
construction, so the reference population is held out from the evaluation in the
ordinary sense. Section~\ref{subsec:main} reports the unlabelled test split as a
control; there the box-annotated images have to be excluded explicitly so that no
evaluation image contributes to its own reference, and even then the readout sees
the test distribution, which a model shipped to a hospital whose population cannot
be sampled in advance could not do.
The anomaly score at position $i$ is the distance from that patch to the same
patch position across the reference population,

\begin{equation}
A(i) \;=\; \min_{r \in \mathcal{R}} \; \big\| f_i(x) - f_i(r) \big\|_2 .
\label{eq:anomaly}
\end{equation}

The comparison is position-wise by construction: a radiograph is roughly
registered by acquisition, so the same index addresses roughly the same anatomy,
and a patch is called unusual relative to what that location normally looks like
rather than relative to the image as a whole. No label of any kind is read, and
$f$ is never updated.

Equation~(\ref{eq:anomaly}) is imported and not introduced: scoring a patch
against the same patch position across a reference set is the standard
construction of industrial anomaly localization, in its nearest-neighbour
\cite{bib-spade}, position-wise distribution \cite{bib-padim} and memory-bank
\cite{bib-patchcore} forms. What is new is not the map but where its supervision
does \emph{not} come from.

That literature also supplies the sharpest objection to this stream, and we state
it rather than answer it. There the reference set is normal by construction,
whereas $\mathcal{R}$ is drawn from the training pool without reading any label,
where 41.5 percent of images carry at least one finding, and the minimum in
Eq.~(\ref{eq:anomaly}) is the operator a single abnormal neighbour damages most.
Bounding that cost would need a bank drawn from the \emph{No Finding} images,
which reads a label and so cannot ship inside the method; we have not run it as a
diagnostic. What can be said without it is narrow: contamination can only lower
the raw anomaly score, but each map is normalised over its own range and
thresholded before a box is taken, so a non-uniform reduction can move a box
either way and \emph{the sign of the effect on localization accuracy is
undetermined}. We report it as an open threat and claim nothing about its
direction.

Because $f$ is frozen, it cannot absorb a preprocessing offset the way a
fine-tuned network does, so each frozen backbone is fed the normalisation
statistics it was pretrained with rather than the classifier's. Which statistics
were used is recorded in every artifact, and Appendix~\ref{app:impl} gives the
values.

\begin{figure*}[!tp]
\centering
\includegraphics[width=\textwidth]{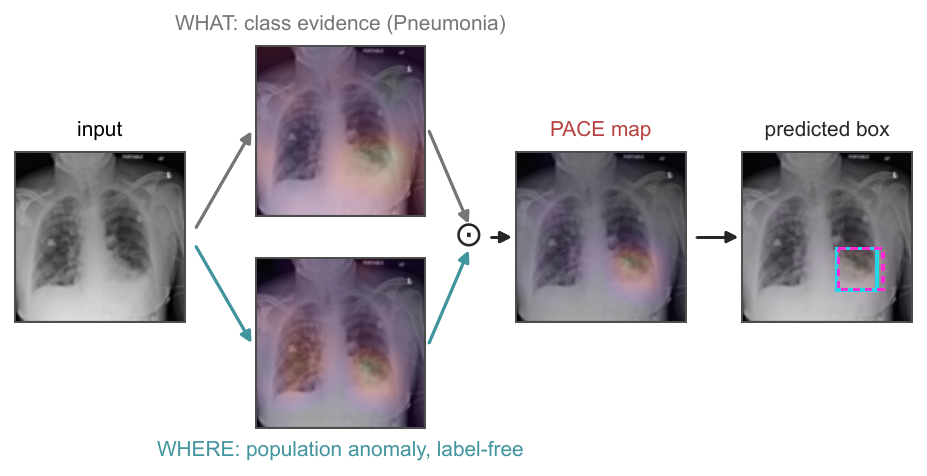}
\caption{PACE combines a label-trained answer to \emph{which finding} with a
label-free answer to \emph{where the image is unusual}. The class-evidence stream
is the class activation map of Eq.~(\ref{eq:cam}), read from the label-trained
classifier. The population-anomaly stream scores each patch by its distance to
the same patch position across a reference set of unlabelled radiographs,
Eq.~(\ref{eq:anomaly}), using a backbone pretrained by masked image modelling on
chest radiographs with no pathology labels. Each map is normalised to $[0,1]$
over its own range and the two are multiplied; the box is the largest connected
component above 0.5 of the product's range. Cyan marks the annotated box and
magenta the predicted one. Every heat map shown is a measured output for one test
image, not an illustration.}
\label{fig:method}
\end{figure*}

\noindent\textbf{Combination and box extraction.}\quad Both maps are bilinearly resampled to the input grid and normalised to $[0,1]$
over their own range, $\widetilde{M}_c$ and $\widetilde{A}$. PACE is their
element-wise product, shown end to end in Fig.~\ref{fig:method},

\begin{equation}
M_c^{\mathrm{PACE}} \;=\; \widetilde{M}_c \;\odot\; \widetilde{A} ,
\label{eq:pace}
\end{equation}

and the predicted box is the bounding box of the largest connected component of
$\{ M_c^{\mathrm{PACE}} > \tau \}$ with $\tau = 0.5$ of the product's own range.
The identical rule at the identical $\tau$ extracts the box from every readout
compared in this paper, including both baselines and each stream taken alone, so
no comparison here is between two box generators.
Section~\ref{subsec:ablation} selects the product against four alternatives:
the mean, the maximum, and two rules that select one stream per image by map
peakedness or by classifier confidence. The maximum, which fires when either
stream is confident, is the only rule that scores below both single streams,
which is what identifies the combination as a conjunction rather than a
disjunction.

\noindent\textbf{Cost.}\quad PACE trains nothing. Building the reference bank is one forward pass of the
frozen backbone over $N$ unlabelled radiographs, done once and reused across all
seeds and all classes. At inference, each image costs one forward pass of the
classifier plus one of the frozen backbone, and Eq.~(\ref{eq:anomaly}) is a
distance against a cached bank. The method therefore attaches to an existing
trained classifier with no change to its weights, its loss, or its training data,
which is also what makes the four-classifier comparison in
Section~\ref{subsec:main} possible.

\section{Localization results}\label{sec:results}

\subsection{PACE against the two baselines}\label{subsec:main}

\begin{figure*}[!tp]
\centering
\includegraphics[width=\textwidth]{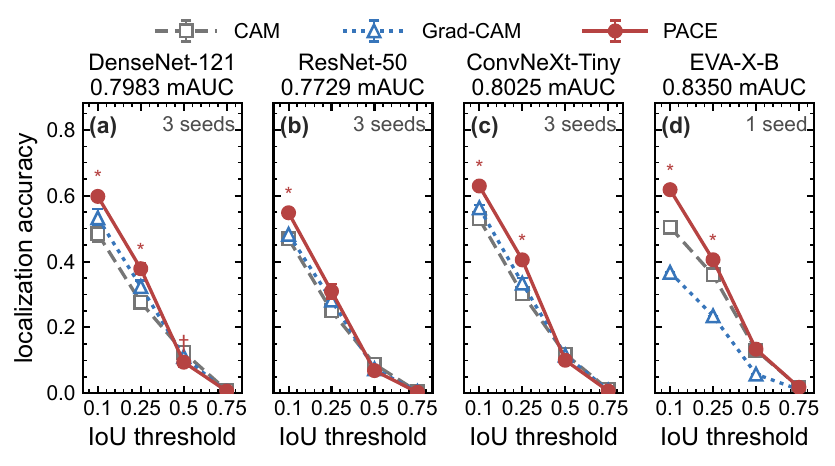}
\caption{PACE against class activation mapping and its gradient-weighted variant
under intersection over union. Panels (a--c) use three classifier seeds;
panel (d) uses one released EVA-X-B checkpoint \cite{bib-evax}. Error bars show
run-to-run variability: classifier seeds for the baselines, classifier seeds and
five reference-population draws for PACE, and only the five draws in panel (d).
They do not sample the 984 boxes. Centres use the designated reference draw.
Stars mark image-clustered intervals excluding zero in every seed and bank cell;
daggers mark agreement across seeds but not all cells. On EVA-X-B,
gradient-weighted mapping is flat on 406 boxes and is not a valid comparison.}
\label{fig:main}
\end{figure*}

Table~\ref{tab:main} and Figure~\ref{fig:main} give PACE against both baselines
across four classifiers on the protected metric. At the shared extraction
threshold the gain over class activation mapping is 0.069 to 0.112 at
$T(\mathrm{IoU}) = 0.1$ and 0.060 to 0.067 over the gradient-weighted variant on
the three classifiers where that readout is not degenerate.
\emph{We state that range once and then qualify it}, because
Section~\ref{subsec:areamatch} shows a shared extraction threshold is not a
shared operating point: at matched box area the same two comparisons are $+0.050$
and $+0.058$. The matched-area figure is the one this paper stands behind, and
the unmatched one is reported because it is what the convention in this
literature would produce.

\begin{table*}[!tp]
\caption{Pooled paired differences over the same 984 boxes under intersection
over union. Upper entries are three-seed means. Lower entries give the weakest
seed and its 95 percent image-clustered interval; clustering is required because
984 boxes lie on 880 images and 93 images carry several boxes. $^{*}$ excludes
zero in every seed and all fifteen classifier-by-bank cells, or all five cells
for the single EVA-X-B checkpoint. $^{\dagger}$ excludes zero in every seed but
not every cell. Unmarked cells are not significant. $T = 0.75$ is omitted because
every compared readout is at or below 0.019. $^{\ddagger}$ compares PACE with its
label-free stream because gradient-weighted mapping is flat on 406 EVA-X-B boxes.}
\label{tab:main}
\small\setlength{\tabcolsep}{1.5pt}
\begin{tabular}{@{}llccc@{}}
\toprule
Classifier & Baseline & $T = 0.1$ & $T = 0.25$ & $T = 0.5$ \\
\midrule
\multirow{4}{*}{DenseNet-121}
  & CAM          & $\mathbf{+0.108}^{*}$ & $\mathbf{+0.105}^{*}$ & $-0.043^{\dagger}$ \\
  &              & \scriptsize $+.076\,[+.048,\,+.104]$ & \scriptsize $+.078\,[+.048,\,+.107]$ & \scriptsize $-.030\,[-.058,\,-.003]$ \\
  & Grad-CAM     & $\mathbf{+0.067}^{*}$ & $\mathbf{+0.057}^{\dagger}$ & $-0.026$ \\
  &              & \scriptsize $+.039\,[+.013,\,+.063]$ & \scriptsize $+.035\,[+.007,\,+.059]$ & \scriptsize $-.014\,[-.039,\,+.011]$ \\
\midrule
\multirow{4}{*}{ResNet-50}
  & CAM          & $\mathbf{+0.069}^{*}$ & $+0.059$ & $-0.026$ \\
  &              & \scriptsize $+.063\,[+.039,\,+.086]$ & \scriptsize $+.025\,[-.004,\,+.055]$ & \scriptsize $-.008\,[-.030,\,+.015]$ \\
  & Grad-CAM     & $\mathbf{+0.060}^{*}$ & $+0.028$ & $-0.009$ \\
  &              & \scriptsize $+.054\,[+.032,\,+.076]$ & \scriptsize $+.018\,[-.009,\,+.047]$ & \scriptsize $-.001\,[-.021,\,+.019]$ \\
\midrule
\multirow{4}{*}{ConvNeXt-Tiny}
  & CAM          & $\mathbf{+0.098}^{*}$ & $\mathbf{+0.111}^{*}$ & $-0.026$ \\
  &              & \scriptsize $+.087\,[+.060,\,+.116]$ & \scriptsize $+.109\,[+.077,\,+.141]$ & \scriptsize $-.012\,[-.039,\,+.013]$ \\
  & Grad-CAM     & $\mathbf{+0.064}^{*}$ & $\mathbf{+0.077}^{*}$ & $-0.022$ \\
  &              & \scriptsize $+.058\,[+.035,\,+.081]$ & \scriptsize $+.075\,[+.046,\,+.106]$ & \scriptsize $-.008\,[-.031,\,+.014]$ \\
\midrule
\multirow{4}{*}{EVA-X-B}
  & CAM          & $\mathbf{+0.112}^{*}$ & $\mathbf{+0.046}^{*}$ & $-0.006$ \\
  &              & \scriptsize $+.112\,[+.082,\,+.140]$ & \scriptsize $+.046\,[+.013,\,+.078]$ & \scriptsize $-.006\,[-.032,\,+.019]$ \\
  & label-free$^{\ddagger}$ & $\mathbf{+0.129}^{*}$ & $\mathbf{+0.109}^{*}$ & $\mathbf{+0.061}^{*}$ \\
  &              & \scriptsize $+.129\,[+.098,\,+.160]$ & \scriptsize $+.109\,[+.075,\,+.142]$ & \scriptsize $+.061\,[+.038,\,+.084]$ \\
\botrule
\end{tabular}
\end{table*}

The gain is significant at the loose thresholds on every classifier and
absent at the strict ones. At $T(\mathrm{IoU}) = 0.5$ the sign reverses in all six
architecture-baseline comparisons on the three classifiers trained here, by
$-0.009$ to $-0.043$, and in the seventh, on the released checkpoint, by $-0.006$;
seven of seven therefore share the sign and the released checkpoint is the
smallest of them rather than an exception. Five of the six are
not significant under clustering, the seventh is not either, and the sixth, against class activation mapping
on DenseNet-121, is a reversal whose interval excludes zero in every seed but in
12 of the 15 cells rather than all of them, so Table~\ref{tab:main} marks it
marginal on the same standard it applies to a gain. The six comparisons share the boxes, the
annotations and the readouts, so they are not independent and six matching signs
are not a formal result; we report the pattern rather than test it. Read with that
caution, it says that PACE finds the right region more often while its smaller box
costs it at strict overlap. Section~\ref{subsec:areamatch} compares the three at
roughly equal box area instead, and the sign there is positive, so the caveat is
tied to the operating point rather than being a fixed property of the readouts;
what that comparison does not establish is which of the two operating points is
the right one to judge strict overlap at. At $T = 0.75$ every readout compared here is at or below 0.019 and no
readout of any kind we ran exceeds 0.025, which is a floor rather than a
measurement, so we draw nothing from that column at all.

Two properties of the reference population in Eq.~(\ref{eq:anomaly}) could each
have produced this gain on their own, and both were tested by running the
alternative rather than argued about. The first is which split it comes from. Ours
is the training pool, held out from the evaluation; drawing it from the unlabelled
test images instead, excluding every box-annotated one, reads no label either but
lets the readout see the test distribution, which makes it transductive and
unavailable to a model shipped to a hospital whose population cannot be sampled in
advance. That control moves the macro by $-0.0073$ to $+0.0050$ over the six
architecture-by-threshold cells, inside the classifier-seed spread in every one,
and every gain at $T = 0.1$ keeps an image-clustered interval excluding zero under
both. The choice of pool is therefore not what the result rests on, and we report
the held-out one.

The second is that the population is one draw of $N = 1000$ images, so we
resampled it four further times at every classifier seed, giving three classifier
seeds by five draws per architecture
(\texttt{scripts/variance\_decomposition.py}). The bank component is 0.0036 to
0.0062 over the six cells and the classifier component 0.0055 to 0.0219, so the
classifier is the larger of the two everywhere here, and the bank still adds
roughly 0.004 to 0.006 that a classifier-seed spread does not see at all.
Combined, the run-to-run variability of a single macro is 0.0066 to 0.0223. Under
the test-pool control the two components are comparable and the bank is the larger
one in the DenseNet-121 cell at $T = 0.1$, so which source dominates depends on
the cell and on the pool; what is stable across both is the size of the bank
component.

That variability is not the paper's main statistical uncertainty, which is
sampling over the 984 boxes, so we checked the conclusions in every one of the
fifteen classifier-by-bank cells rather than in the mean alone. At $T = 0.1$ the
image-clustered interval excludes zero in all 15 cells of all six comparisons,
with the smallest lower bound at $+0.0123$. At $T = 0.25$ it does not: 15 of 15
for both ConvNeXt-Tiny comparisons and for DenseNet-121 against class activation
mapping, 13 of 15 against gradient-weighted mapping, and 11 and 10 of 15 for
ResNet-50, whose two cells Table~\ref{tab:main} already reports as not significant.
So the loose-threshold result survives every draw of the population and every
seed, while the DenseNet-121 gain at $T = 0.25$ against gradient-weighted mapping
is marginal, and Table~\ref{tab:main} marks it as marginal rather than as
significant.

Treating the boxes as independent would have overstated the evidence, which is why
none of the marks above come from a box-level test. Resampling images instead of
boxes, so that every box of a drawn image travels with it, costs the two ResNet-50
comparisons at $T = 0.25$ their significance and changes nothing at $T = 0.1$. A
paired McNemar test over the boxes puts those two ResNet-50 cells at $p = 0.12$
and $p = 0.24$ anyway, so no claim in this paper ever depended on the
anti-conservative version. That the correction is small is expected rather than
lucky: 787 of the 880 images carry exactly one box.

The fourth classifier tests whether the result depends on weak classifiers.
PACE was developed on models at 0.7729 to 0.8025 while the comparison methods run
0.808 to 0.835, so the gain could have been a property of a weak backbone. We
therefore ran the identical readout, controls and paired test on a released
foundation-model checkpoint \cite{bib-evax}, reached through an adapter to the
existing evaluation path. Class activation mapping is exact for
that architecture too: it pools patch tokens, applies a layer normalisation and
then one linear layer, so once the normalisation's two scalars are computed from
the pooled vector the class score decomposes into an exact per-token map. The
adapter asserts this numerically at run time before producing any result, and
aborts otherwise; on the test split it reconstructs the network's own logits to a
maximum absolute error of $1.4 \times 10^{-6}$ on a logit scale of 7.9. That
assertion guards the pipeline rather than being a reported measurement, and it
leaves no artifact of its own. The same threshold dependence appears on a
classifier whose mean area under the curve is 0.037 higher.

That row has three limitations. There is one
released checkpoint, so it carries no classifier-seed spread; its run-to-run bar is
the reference population alone and its statistical uncertainty is the paired test
over the 984 boxes. Gradient-weighted mapping is degenerate on this transformer,
returning a flat map on 406 of 984 boxes, so class activation mapping is the
meaningful baseline there. And because EVA-X-B is fine-tuned from the same
masked-image-modelling checkpoint we use as the label-free stream, that pairing
shares an initialisation; repeating it with a backbone that shares no weights
with the classifier \cite{bib-xiao2023} still gives a significant gain at
$T = 0.1$ ($+0.0661$, $p = 2.3 \times 10^{-6}$), at about three fifths of the size.

\subsection{The second corpus}\label{subsec:chexloc_results}

Table~\ref{tab:chexlocmatched} repeats the comparison on CheXlocalize at matched
box area, since Section~\ref{subsec:areamatch} showed that a shared extraction
threshold is not a shared operating point on this corpus either.

\begin{table*}[!tp]
\caption{PACE on CheXlocalize at matched box area, three classifier seeds, boxes
recalibrated so all readouts emit $0.70\times$ to $0.88\times$ the target area.
Intervals are image-clustered, which matters far more here than on ChestX-ray14:
2,744 boxes fall on 584 images, 4.70 per image with only 17.5 percent
singletons, against 1.12 and 89.4 percent there. $\Delta$ is the mean over three classifier seeds. `Clustered' is whether the
image-clustered interval excludes zero in the governing seed, which is the
significance standard used throughout this paper; the interval itself is a
different estimator from the mean and is therefore not printed beside it. The
last column counts how many of the fifteen classifier-by-reference-bank cells
have an interval excluding zero.}\label{tab:chexlocmatched}
\begin{tabular}{llccc}
\toprule
Comparison & $T(\mathrm{IoU})$ & $\Delta$, 3 seeds & clustered & cells \\
\midrule
\multirow{3}{*}{over class activation}
  & 0.1 & $+0.013 \pm 0.005$ & no & 6 of 15 \\
  & 0.25 & $\mathbf{+0.075} \pm 0.010$ & \textbf{yes} & \textbf{15 of 15} \\
  & 0.5 & $\mathbf{+0.028} \pm 0.009$ & \textbf{yes} & \textbf{15 of 15} \\
\midrule
\multirow{3}{*}{over gradient-weighted}
  & 0.1 & $\mathbf{+0.117} \pm 0.010$ & \textbf{yes} & \textbf{15 of 15} \\
  & 0.25 & $\mathbf{+0.109} \pm 0.014$ & \textbf{yes} & \textbf{15 of 15} \\
  & 0.5 & $+0.018 \pm 0.010$ & no & 10 of 15 \\
\midrule
\multirow{3}{*}{over the label-derived prior}
  & 0.1 & $+0.048 \pm 0.038$ & no & 10 of 15 \\
  & 0.25 & $\mathbf{+0.143} \pm 0.019$ & \textbf{yes} & \textbf{15 of 15} \\
  & 0.5 & $\mathbf{+0.047} \pm 0.003$ & \textbf{yes} & \textbf{15 of 15} \\
\bottomrule
\end{tabular}
\end{table*}

\textbf{The gain is real, smaller than on ChestX-ray14, and lives at the strict
thresholds.} It survives every one of the fifteen classifier-by-bank cells at
$T(\mathrm{IoU}) = 0.25$ and $0.5$ against all three comparators, and does not
survive at $0.1$ against class activation mapping. That the gain sits at strict
overlap here and at loose overlap on ChestX-ray14 follows directly from the box
areas in Section~\ref{subsec:areamatch} and is not a separate finding.

\textbf{The head-to-head that the novelty claim needs holds on both corpora.}
Replacing the label-free second map with a label-derived one, entering the same
product by the same rule, costs $+0.143$ at $T(\mathrm{IoU}) = 0.25$ and
$+0.047$ at $0.5$, at matched area, in every cell. \emph{Having} a second map is
therefore not what carries the result. Which \emph{property} of the second map
does is bounded rather than isolated by this substitution, because the two maps
differ in resolution as well as in supervision, as
Section~\ref{subsec:secondmap} states.

\subsection{Where the second map comes from}\label{subsec:secondmap}

Section~\ref{sec:related} places this paper's only novelty claim on the
\emph{source} of the second spatial map rather than on having one, and until now
that contrast was argued rather than measured: the ablation removes the second
stream, which tests whether a second map helps, not where it should come from.
We therefore build the label-derived counterpart of the same shape. Following
\cite{bib-jung2023}, for every class we average the class activation maps of every
training image labelled positive for it, each normalised over its own range, into
a per-class occurrence prior, and multiply that into the same class-evidence
stream by the same rule. The prior is built on the training split only, reads
labels by construction, and needs no retraining. It is not a reimplementation of
\cite{bib-jung2023}, which trains a network under its prior; it is the
label-derived analogue of PACE's second stream, which is the comparison the claim
requires.

The two second maps differ in shape as well as in source, and both differences
matter. The population-anomaly map is one map per image and carries no class; the
occurrence prior is one map per class and carries no image. Both enter the same
$\odot$ of Eq.~(\ref{eq:pace}).

\begin{table*}[!tp]
\caption{The second map's source, at matched box area. DenseNet-121, three seeds,
intersection over union, pooled paired differences over the 984 boxes; the upper
number is the three-seed mean and the bracket is the governing seed's 95 percent
image-clustered interval, on the convention of Table~\ref{tab:areamatch}. In the
second row the two lines differ in sign at $T = 0.25$, which is what a
non-significant cell looks like when the seeds straddle zero and the governing
seed is therefore chosen without reference to direction. The occurrence prior is
taken at $\tau = 0.60$, where its box is 1.54 times the annotation against PACE's
1.40, and class activation mapping at $\tau = 0.75$ at 1.45 times. The first row
is the comparison the novelty claim needs; the second is why it has to be made at
matched area.}
\label{tab:secondmap}
\small\setlength{\tabcolsep}{3pt}
\begin{tabular}{lccc}
\toprule
 & $T = 0.1$ & $T = 0.25$ & $T = 0.5$ \\
\midrule
PACE over the occurrence prior
  & $\mathbf{+0.064}$ & $\mathbf{+0.090}$ & $\mathbf{+0.063}$ \\
  & \scriptsize $[+.026,+.075]$ & \scriptsize $[+.024,+.080]$ & \scriptsize $[+.033,+.070]$ \\
\midrule
Occurrence prior over CAM
  & $-0.014$ & $-0.040$ & $-0.025$ \\
  & \scriptsize $[-.035,+.010]$ & \scriptsize $[-.013,+.035]$ & \scriptsize $[-.034,-.003]$ \\
\botrule
\end{tabular}
\end{table*}

\textbf{A label-derived second map buys nothing here once its box area is
matched.} At the fixed $\tau = 0.5$ the occurrence prior appears to raise
accuracy over class activation mapping alone by $+0.049$ at
$T(\mathrm{IoU}) = 0.1$, which would be a substantial effect. It is a box-size
effect: the prior's box is 2.55 times the annotation and the class activation
map's is 5.92, and once both are brought to about 1.5 times the difference is
$-0.014$ with an interval containing zero, and negative at $T = 0.5$. The
label-free second map, at the same matched area, is worth $+0.050$ to $+0.058$
over the same baselines, and PACE beats the label-derived prior directly by
$+0.064$ to $+0.090$, significant at every threshold and in every seed.

We state what that does and does not license. It is evidence about \emph{these two
concrete second maps} and not about label-derived maps as a class: one coarse,
post-hoc, image-independent prior failed here, and a different label-derived
construction, in particular one that varies with the image or is learned rather
than averaged, could behave otherwise. What it does replace is a claim this paper
had been making in prose alone. The direction it points is the one
Section~\ref{sec:intro} predicts from the data-generating process, and the paper
now has one measurement of it where it had none.

Two limits on that reading. The prior is a coarse map by construction, its box
covering 8.9 times the annotation when used alone, so part of what separates the
two sources is resolution and not only supervision; and this substitution keeps
everything else fixed but does not test \cite{bib-jung2023}'s own method, which
trains under its prior and may recover what the post-hoc form does not.

\subsection{Ablation}\label{subsec:ablation}

\begin{figure*}[!tp]
\centering
\includegraphics[width=\textwidth]{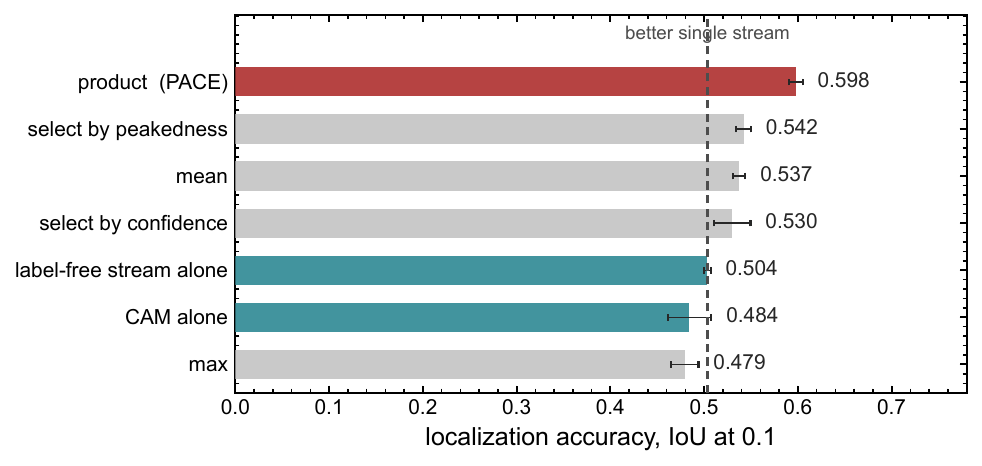}
\caption{Combination-rule ablation on DenseNet-121. Bar centres are class macros
over eight findings and 984 boxes on the designated reference draw, matching
Table~\ref{tab:ablation}. Error bars combine three classifier seeds and five bank
draws, except where a stream does not use one source of variation; they exclude
sampling over boxes. The dashed line marks the better single stream. Only the
product clears it by a margin, while the maximum falls below both streams.}
\label{fig:rule}
\end{figure*}

Table~\ref{tab:ablation} removes each component in turn and replaces the
combination rule with each alternative. Figure~\ref{fig:rule} plots the rules
against the better single stream.

\begin{table*}[!tp]
\caption{DenseNet-121 ablation on 984 boxes under intersection over union. The
upper block removes one stream; the lower replaces the product in
Eq.~(\ref{eq:pace}). $s_{\mathrm{clf}}$ is the spread over three classifier seeds,
$s_{\mathrm{bank}}$ the spread over five draws of the 1000-image reference
population, and $s_{\mathrm{total}}$ their quadrature sum. These are rerun
variabilities, not sampling errors over boxes. Class activation mapping has no
bank component, while the label-free stream has no classifier component.}
\label{tab:ablation}
\begin{tabular}{lccccc}
\toprule
Variant & $T(\mathrm{IoU}) = 0.1$ & $s_{\mathrm{clf}}$ & $s_{\mathrm{bank}}$
        & $s_{\mathrm{total}}$ & Change \\
\midrule
\textbf{PACE (full)}                & \textbf{0.5978} & 0.0064 & 0.0036 & \textbf{0.0073} & \\
w/o population-anomaly stream & 0.4844 & 0.0233 & n/a & 0.0233 & $-0.1133$ \\
w/o class-evidence stream    & 0.5036 & 0.0000 & 0.0033 & 0.0033 & $-0.0941$ \\
\midrule
$\odot \rightarrow$ select by map peakedness & 0.5421 & 0.0075 & 0.0024 & 0.0078 & $-0.0557$ \\
$\odot \rightarrow$ mean & 0.5372 & 0.0057 & 0.0038 & 0.0068 & $-0.0605$ \\
$\odot \rightarrow$ select by confidence & 0.5301 & 0.0193 & 0.0011 & 0.0193 & $-0.0676$ \\
$\odot \rightarrow$ maximum & 0.4794 & 0.0142 & 0.0031 & 0.0145 & $-0.1183$ \\
\botrule
\end{tabular}
\footnotetext{Point estimates are class macros over the eight annotated findings,
on the designated draw of the held-out reference population. They are computed
from the same per-box records as the paired tests of Table~\ref{tab:main} and on
the same bank, but they are not the same aggregate: those tests are pooled over
the 984 boxes, so a difference of two rows here is not a paired difference there
and the two are not expected to agree. The macro difference between the first two
rows is 0.1133 and the corresponding pooled paired difference in
Table~\ref{tab:main} is 0.108. The mean over all five draws differs by at most
0.0048 on any row and is inside that row's $s_{\mathrm{total}}$ in every case. Every value
is the same metric at the same threshold, and the ordering of the rules is
identical under intersection over the predicted box, where the corresponding
values are 0.6098, 0.5496, 0.5407, 0.5374 and 0.4833.}
\end{table*}

Neither stream carries the result alone, and the two selection rules, which pick
one stream per image rather than combining them, both sit between the product and
the streams. The decisive row is the maximum: a disjunction of the two streams
scores below either of them, so the evidence they carry is conjunctive.

\section{Selection, and two tests of it}\label{sec:confirm}

ChestX-ray14 releases its boxes only inside the test split, so this benchmark has
no localization validation set. Four choices behind the readout of
Section~\ref{sec:method} were therefore made against the same 984 boxes
Section~\ref{subsec:main} reports on: the frozen backbone, the rule that combines
the two maps, the extraction threshold, and the mask-to-box rule. Three classifier
seeds, five reference-bank draws, Holm corrections and image-clustered intervals
all quantify variability \emph{after} those choices; none of them restores an
independent test set, and checking a test set under several configurations is
itself a use of localization supervision \cite{bib-choe2020}. The paper's own
sensitivity analyses show the choices are load-bearing rather than incidental:
Section~\ref{subsec:boxrule} moves this readout by 0.216 by changing the box rule
alone. This section reports what happens when the selection is redone under a protocol
that separates the cases used to choose from the cases used to score, and it does
so twice, once inside ChestX-ray14 and once on the second corpus. \emph{Neither is
a held-out test in the usual sense}, and Section~\ref{subsec:notestablished} says
what each does and does not support.

\subsection{A post-hoc stability split inside ChestX-ray14}\label{subsec:heldout}

The 880 box-annotated images are partitioned at \emph{patient} level into halves,
frozen before anything below is computed; the halves share no patient and no
image, and hold 491 and 493 boxes over 440 images each. The whole selection is
then rerun on the selection half, by the same criterion used to make the choices
originally, the pooled accuracy of the fused readout at the headline
operating point. Closing the backbone axis required six new localization passes,
the product readout on the medical\_mae and DINOv2 reference banks at three
classifier seeds each; every other candidate was already on disk. The winner is
then evaluated on the other half, once after this partition was drawn.
\emph{That half was never withheld}: every image in it had already contributed to
developing the readout, to choosing the reported configuration and to enumerating the
candidates, and no partition drawn afterwards changes that. What the split
measures is stability, not confirmation. Intervals in this section resample
\emph{patients} rather than images, because the partition is at patient level and
55 of the 363 evaluation-half patients contribute more than one image; the effect of that
correction is to widen the intervals by about 0.003, and it changes no verdict.

\emph{Two properties of the candidate set are declared here rather than left for a
reader to infer.} It was enumerated after the answer on the full 984 boxes was
known, so the set itself is a degree of freedom this test does not examine. And it
varies one factor at a time around the reported configuration rather than
searching the four axes jointly, so its 20 labels resolve to 17 distinct
evaluations and the four that tie at the top are one configuration seen from four
directions, not four independent wins.

\begin{table*}[!tp]
\caption{Selection repeated on a post-hoc patient-level stability split. Neither
half is held out in the usual sense because both shaped the readout before the
partition. Left: pooled $\mathrm{IoU}@0.1$ on the selection half, averaged over
three seeds; 20 labels give 17 distinct evaluations. Right: the selected
configuration on 493 boxes over 440 evaluation-half images. Intervals are 95
percent patient-clustered intervals in the governing seed because 55 of 363
patients contribute several images. The table measures stability, not
confirmation.}\label{tab:heldout}
\begin{tabular}{lc@{\hskip 2em}lc}
\toprule
Candidate, selection half & acc. & Evaluation half & value \\
\midrule
\textbf{backbone = EVA-X} & \textbf{0.6198} & selected on selection half & EVA-X, product, \\
\textbf{rule = product}   & \textbf{0.6198} &                 & $\tau = 0.5$, largest \\
\textbf{threshold $\tau = 0.5$} & \textbf{0.6198} & & \\
\textbf{box rule = largest} & \textbf{0.6198} & over class activation & $\mathbf{+0.1285}$ \\
rule = select-by-peak & 0.5798 & \quad governing interval & $[+0.054, +0.133]$ \\
rule = mean & 0.5750 & over gradient-weighted & $\mathbf{+0.0859}$ \\
rule = select-by-confidence & 0.5750 & \quad governing interval & $[+0.028, +0.100]$ \\
backbone = medical\_mae & 0.5703 & & \\
rule = max & 0.5227 & redrawing the partition & \\
box rule = union & 0.4229 & \quad selection, 5 splits & unchanged \\
backbone = DINOv2 & 0.4189 & \quad over class activation & $+0.091$ to $+0.137$ \\
box rule = peak & 0.4046 & \quad splits significant & 5 of 5 \\
$\tau = 0.55$ & 0.4067 & & \\
$\tau = 0.6$ & 0.3802 & selected candidate, absolute & \\
$\tau = 0.9$ & 0.0821 & \quad selection $\to$ evaluation & $0.6198 \to 0.6518$ \\
\bottomrule
\end{tabular}
\end{table*}

\textbf{The selection half selects the configuration this paper reports, and the
evaluation half reproduces the gain} (Table~\ref{tab:heldout}). On 493 boxes it
reads $+0.1285$ over class activation mapping and $+0.0859$ over the
gradient-weighted variant at $T(\mathrm{IoU}) = 0.1$, with a patient-clustered
interval excluding zero in every classifier seed. Redrawing the partition four
further times changes neither the selected configuration nor the sign, with the
margin over class activation mapping running from $+0.091$ to $+0.137$ and an
interval excluding zero in all five. That sweep is a sensitivity check on an
arbitrary partition and not five independent confirmations, since all five
outcomes were inspected.

\emph{Selection inflation is tested on the quantity the selection maximised},
which is the candidate's absolute pooled accuracy and never its margin over a
baseline. Maximising an accuracy on a finite sample inflates it, so the selected
candidate should score \emph{lower} out of sample. It does not: $0.6198$ on the
selection half against $0.6518$ on the evaluation half, a shift of $-0.032$ in the
direction opposite to inflation, while the two baselines, which nothing selected,
move $+0.008$ and $+0.006$ the ordinary way. \emph{Across the five partitions the
shift runs from $-0.039$ to $+0.004$, so it is negative in four and marginally
positive in one}, and reporting only the first would be quoting one draw of an
arbitrary partition.

Two limits on how far that reading goes. The halves are not exchangeable in
difficulty, and the observation that our readout gains while both baselines lose
is as consistent with a partition that happens to suit one and not the others as
it is with an absence of inflation, so this is weak evidence in the right
direction and not a proof. And it is reported because it is the check the
selection criterion licenses: a comparison of the two halves' \emph{differences},
which an earlier version of this section made and which is why they are not
tabulated, tests a quantity nothing here optimised.

The test can go red, which was checked before it was believed. Had the
selection half chosen any of the losing candidates, the evaluation half would
report a loss against class activation mapping: $-0.008$ for the maximum rule,
$-0.070$ for the DINOv2 bank, $-0.439$ at $\tau = 0.9$.

\subsection{The frozen configuration on a second corpus}\label{subsec:frozen}

The second test is external and its verdict is split. Taking the configuration
above unchanged, with no recalibration of any kind, and applying it once to
CheXlocalize gives, at $T(\mathrm{IoU}) = 0.1$ over 2,744 boxes,
$\mathbf{-0.056}$ against class activation mapping with an image-clustered
interval excluding zero, and $\mathbf{+0.042}$ against the gradient-weighted
variant, also excluding zero; at $0.25$ the two read $+0.006$, not significant,
and $+0.026$, significant; at $0.5$ and $0.75$ neither separates.
\textbf{Transferred without recalibration, the readout beats one baseline and
loses to the other.} Section~\ref{subsec:areamatch} identifies the mechanism, and
identifying it does not undo it: the shared extraction threshold is not a shared
operating point, and on this corpus the ordering of box areas is reversed, so at
the loosest threshold a merely containing box scores for free. The matched-area
comparison in Table~\ref{tab:chexlocmatched} answers a different and useful
question, but it recalibrates against the target boxes, so it is exploratory in a
sense this transfer is not: nothing in the transfer was chosen after seeing
CheXlocalize. That is the one respect in which the external test is stronger than
the internal one, whose candidate set was not fixed in advance either. Neither is
a confirmatory test in the sense \cite{bib-choe2020} sets out, which would need
the whole procedure frozen before any box was scored.

\subsection{What neither test establishes}\label{subsec:notestablished}

Both tests are post-hoc, and what follows from them is a stability claim rather
than a causal one. The answer on the full 984 boxes was known before either
partition was drawn, the candidate set was enumerated with that answer in hand,
and no partition drawn afterwards unsees either. So the internal test says the
result is stable under a patient-level partition, on the axes tested, in a
candidate set built one factor at a time; it does not establish that the four
choices are innocent of it, which would need a procedure fixed before the answer
was seen. The external test says the operating point does not transfer, which is
the stronger of the two statements because nothing about it was chosen here.

One thing remains outside the reach of any partition of this data. The readout
itself, the idea of multiplying a class-evidence map by a population-anomaly map,
was arrived at while looking at these boxes. A post-hoc stability split constrains
the choices made within a hypothesis; it does not make the hypothesis
pre-registered. The measurements of Sections~\ref{subsec:rq1} to \ref{subsec:analysis} do not
carry this liability, for the reason given in Section~\ref{sec:intro}.

\section{What a comparable localization number has to state}\label{sec:protocol}

Five of this paper's measurements are not about any method. They are about
choices that published numbers on this benchmark depend on and mostly do not
report, each found because a number moved when the choice changed. The size of
each effect is measured in the section named beside it; collected here, they are
what a reader needs before two numbers can be compared.

\begin{enumerate}
\item \textbf{The results regime.} Two bodies of classification results on this
benchmark do not cite each other and the second starts above where the first ends
(Section~\ref{sec:related}). State which one a number belongs to.

\item \textbf{The box-extraction threshold.} Sweeping it moves pooled accuracy at
$T(\mathrm{IoU}) = 0.1$ by 0.236 to 0.553 depending on the readout, and the
conventional $\tau = 0.5$ is the best setting for neither baseline
(Table~\ref{tab:sweep}). State $\tau$, and prefer a comparison at matched box area
to one at matched $\tau$.

\item \textbf{The mask-to-box rule.} Almost nobody states it. It moves the two
baselines by 0.021 and 0.012 and a peaked readout by 0.216, and the sensitivity
is predicted by how fragmented the thresholded mask is
(Section~\ref{subsec:boxrule}). State the rule.

\item \textbf{A position null.} A macro without one does not separate grounding
from a position prior, and against size-matched controls the standard readouts
clear the null on two to four of seven informative findings
(Table~\ref{tab:priorall}). The practice is imported from natural-image
evaluation \cite{bib-choe2020} and costs one extra pass.

\item \textbf{Whether the design choices were made on the reported boxes.}
ChestX-ray14 releases no localization validation set, so any method tuned on this
benchmark was tuned on its own test set. State it, and if the boxes allow,
partition them at patient level and evaluate the selected configuration on the
other half (Section~\ref{sec:confirm}). Call that a stability analysis and not a
held-out test: the cases were inspected before the partition existed.
\end{enumerate}

Two smaller items follow from the same reasoning. Intersection over the predicted
box divides by the predicted area alone, so it rewards a smaller box for free and
should not carry a between-method claim on its own; and the released box file
holds 984 boxes over 880 images, so a table quoting different counts is measuring
something else, as Wang et al.'s own appendix does at 1,600 boxes over 983 images
\cite{bib-wang2017}.

We do not claim these five are exhaustive. They are the ones that changed a
number in this paper, and three of them changed one of ours by more than the
effect we set out to measure.

\section{Conclusion}\label{sec:conclusion}

Weakly supervised localization scores on chest radiographs contain a large
position prior. Size-matched controls leave the standard readouts above the null
on only two to four of seven informative findings on ChestX-ray14 and two of
eight on CheXlocalize. The test is not impossibly strict: a benchmark drawn by
three board-certified radiologists clears the same null on four to seven
findings. A macro score alone therefore does not establish spatial grounding.

The reported value also depends on the results regime, extraction threshold and
mask-to-box rule. Predicted box area can reverse a paired comparison, and the
direction of that reversal differs between corpora. These choices must accompany
the score.

PACE exposed those dependencies. \emph{In an exploratory matched-area analysis},
its label-free population-anomaly map improves strict-overlap accuracy on both
corpora and beats a label-derived map under the same product rule. That analysis
is exploratory on either corpus for the same reason: the extraction thresholds
that define the matched operating point were themselves chosen against the boxes
the comparison is scored on. It also clears the position null
on more findings than the two standard readouts. The gain is conditional: the
largest-connected-component rule moves PACE by 0.216, PACE is the least faithful
readout to the classifier's occlusion map, and every design choice was made after
the ChestX-ray14 boxes had been inspected. \emph{One threat stays open}: PACE
emits the smallest box of the three, so its advantage at loose thresholds may be
concentration rather than relocation, and the matched-area analysis that
addresses it sets its thresholds against the same boxes. The patient-level split can test only
post-hoc stability. Frozen transfer to CheXlocalize is the clean test, and it is
mixed: PACE beats the gradient-weighted variant and loses to class activation
mapping. PACE is therefore not established as a generally better localizer.

The remaining limits are concrete. The two frozen backbones differ in objective,
corpus and training recipe, so these experiments do not isolate pretraining
domain. ChestX-ray14 provides 984 boxes over 8 of 14 findings, while
CheXlocalize provides 2,744 boxes over 10 findings through converted radiologist
segmentations. The human ceiling compares final masks between reader groups but
cannot measure agreement within either group. A confirmatory test needs a box
corpus whose final annotations are not inspected until the readout and its box
rule have been fixed.

\begin{appendices}

\section{Classification: reproduction and comparison}\label{subsec:repro}

Before measuring anything new we reproduce a published baseline matched on both
architecture and split. Table~\ref{tab:cls} gives per-class and mean values for
the three classifiers over three seeds. DenseNet-121 reaches
$0.7983 \pm 0.0009$. That is 0.0067 to 0.0087 below the 0.805 to 0.807 that two
independent references report under different test protocols
\cite{bib-ranjan2018,bib-wangh2021}, so it sits inside this project's $\pm 0.01$
reproduction tolerance rather than inside their band, and the recipe here is
deliberately not matched to either.

We also reproduce the five-rung augmentation
ladder of \cite{bib-ranjan2018} at three seeds each under its own recipe,
including its unweighted loss. All five rungs land inside a $\pm 0.01$ tolerance
on the three-seed mean, and the shape reproduces in that all the gain arrives in
the first two rungs. Two deviations are recorded rather than tuned away.
Every rung undershoots its own published value, by 0.0016 to 0.0094,
which is a systematic offset rather than seed noise; and the published
monotonicity does not reproduce, since our fourth rung sits 0.0027 below the
third where the source reports a gain of 0.0002. At our measured seed spreads of
0.0007 to 0.0013, that source's differences among its top three rungs are not
distinguishable from zero, which is a statement its own single-run protocol
cannot make.

\begin{table*}[!tp]
\caption{Classification on the official ChestX-ray14 test split of 25,596 images,
single crop, mean and standard deviation over three seeds. All values are in the
official-split regime defined in Section~\ref{subsec:data}.}
\label{tab:cls}
\begin{tabular}{lccc}
\toprule
Finding & DenseNet-121 & ResNet-50 & ConvNeXt-Tiny \\
\midrule
Atelectasis        & 0.7581 (50) & 0.7405 (64) & 0.7640 (10) \\
Cardiomegaly       & 0.8717 (11) & 0.8406 (89) & 0.8848 (85) \\
Effusion           & 0.8170 (21) & 0.7991 (36) & 0.8200 (23) \\
Infiltration       & 0.6823 (49) & 0.6851 (9)  & 0.6997 (33) \\
Mass               & 0.7916 (34) & 0.7417 (92) & 0.8144 (14) \\
Nodule             & 0.7341 (98) & 0.6982 (22) & 0.7532 (28) \\
Pneumonia          & 0.7012 (63) & 0.6941 (75) & 0.7162 (19) \\
Pneumothorax       & 0.8495 (55) & 0.8293 (99) & 0.8526 (68) \\
Consolidation      & 0.7344 (17) & 0.7140 (28) & 0.7408 (25) \\
Edema              & 0.8304 (46) & 0.8221 (2)  & 0.8434 (28) \\
Emphysema          & 0.9035 (41) & 0.8500 (68) & 0.8945 (52) \\
Fibrosis           & 0.8231 (54) & 0.7928 (119)& 0.8145 (116) \\
Pleural thickening & 0.7640 (42) & 0.7373 (24) & 0.7624 (58) \\
Hernia             & 0.9149 (106)& 0.8759 (157)& 0.8746 (299) \\
\midrule
\textbf{Mean}      & \textbf{0.7983 (9)} & 0.7729 (25) & \textbf{0.8025 (31)} \\
\botrule
\end{tabular}
\footnotetext{Parenthesised digits are the standard deviation over three seeds in
units of the last decimal, so 0.7983 (9) means $0.7983 \pm 0.0009$.}
\end{table*}

\section{Implementation}\label{app:impl}

All experiments run on NVIDIA RTX 3090 (24\,GB) cards. The committed runs record
two hosts of one cluster, both carrying four such cards under Ubuntu 20.04 with
driver 535.183.01, with PyTorch 2.5.1
(CUDA 12.1), torchvision 0.20.1, timm 1.0.11, NumPy 1.26.4 and SciPy 1.17.1,
pinned to identical versions on every node. Configurations are parallelised
across nodes rather than each model across GPUs, so no run uses more than one
card. Radiographs are
decoded once into a fixed-size unsigned 8-bit memory-mapped cache at 224 pixels
and never re-decoded inside the training loop.

Classifiers are DenseNet-121 \cite{bib-densenet}, ResNet-50 \cite{bib-resnet} and
ConvNeXt-Tiny \cite{bib-convnext}, each initialised from ImageNet
\cite{bib-imagenet}, with the grayscale channel replicated three times. Training
uses Adam \cite{bib-adam} at a learning rate of $10^{-4}$, no weight decay, batch size 64, up to
40 epochs with early stopping at patience 8 on the validation mean area under the
curve, mixed precision, and a horizontal flip as the only augmentation. Model
selection monitors the validation metric and never the test split. Seeds are
recorded, and every run writes its resolved configuration, its commit hash and
its metrics to its own directory.

The frozen backbone for the population-anomaly stream is the base variant of a
masked-image-modelling model pretrained on 520k radiographs \cite{bib-evax}; two
alternatives are reported for contrast, a masked autoencoder \cite{bib-mae} pretrained on 266k
radiographs \cite{bib-xiao2023} and a self-supervised vision transformer
\cite{bib-vit} pretrained on 142M natural images \cite{bib-dinov2}. The last is used in its register variant, since
without registers its class-token attention collapses onto high-norm background
tokens \cite{bib-registers} and the readout degenerates to a fixed corner box.
Each frozen backbone receives its own pretraining normalisation, read from that
project's own transform code rather than assumed: 0.4919 / 0.2851 for the
radiograph model of \cite{bib-evax}, 0.5056 / 0.2520 for that of
\cite{bib-xiao2023}, and the channel-wise ImageNet statistics for
\cite{bib-dinov2}. The reference bank holds $N = 1000$ unlabelled radiographs
drawn from the training pool, which contains no box-annotated image; the
test-split draw, which does and from which they are excluded, is the transductive
control of Section~\ref{subsec:main}.

\section{Equivalence-test bookkeeping}\label{app:tost}

Two comparisons by two overlap measures by four thresholds gives sixteen cells.
Three are excluded as floor effects, where both arms sit below 0.10 accuracy and
an interval narrower than the margin says only that neither readout can do the
task, leaving thirteen testable, of which two support equivalence. Restricting to
the protected metric leaves eight cells, two excluded at $T = 0.75$ as floor
effects and six testable, of which one supports equivalence. The second of the
two is on intersection over the predicted box at $T = 0.5$, which the box size
confounds in the label-free readout's favour, so it is not evidence and is not
counted as such; the supportable count is one of six.

\section{Comparison with published methods}\label{app:compare}

Rather than quote a table, we scored eight released checkpoints on our own frozen
test manifest with our own metric code, each under its source project's eval
transform and normalisation. Table~\ref{tab:compare} reports both our value and
the published one. Every one of the eight lands within 0.0013 of its
published mean, and for the strongest of them, EVA-X-B, the reproduction and the
published value agree exactly. That validates the evaluation code before any row
derived from it is used.

\begin{table*}[!tp]
\caption{Comparison on the official ChestX-ray14 test split, in three blocks that
are not on the same footing. The first is pretrained on hundreds of thousands of
radiographs or image-report pairs, and we do not claim to beat it. The second is
CIPL, which like ours starts from ImageNet but trains a prototype framework at
$512 \times 512$; its number is quoted from the paper and is the only row here we
did not run. The third is ours, from ImageNet with 77,919 training images.
Reproduced values are scored on our manifest with our metric code under each
source project's own eval transform.}
\label{tab:compare}
\begin{tabular}{llccc}
\toprule
Method & Backbone & Pretraining data & Reproduced & Published \\
\midrule
EVA-X-B \cite{bib-evax}      & ViT-B/16     & 520k radiographs   & \textbf{0.8350} & 0.835 \\
EVA-X-S \cite{bib-evax}      & ViT-S/16     & 520k radiographs   & 0.8332 & 0.833 \\
EVA-X-Ti \cite{bib-evax}     & ViT-Ti/16    & 520k radiographs   & 0.8227 & 0.824 \\
Medical MAE \cite{bib-xiao2023} & DenseNet-121 & 266k radiographs & 0.8131 & 0.813 \\
MGCA \cite{bib-mgca}         & ResNet-50    & image-report pairs & 0.8128 & 0.813 \\
MedKLIP \cite{bib-medklip}   & ResNet-50    & image-report pairs & 0.8115 & 0.812 \\
BioViL \cite{bib-biovil}     & ResNet-50    & image-report pairs & 0.8082 & 0.808 \\
MoCo v2 \cite{bib-mocov2}    & DenseNet-121 & radiographs        & 0.8081 & 0.808 \\
\midrule
CIPL \cite{bib-cipl}         & DenseNet-121 & ImageNet           & not re-run & 0.828 \\
\midrule
Ours                         & ConvNeXt-Tiny & ImageNet          & 0.8025 (31) & \\
Ours                         & DenseNet-121  & ImageNet          & 0.7983 (9)  & \\
Ours                         & ResNet-50     & ImageNet          & 0.7729 (25) & \\
\botrule
\end{tabular}
\footnotetext{Reproduced rows in the upper block are single released checkpoints
and have no seed spread. Rows for our models are the mean over three seeds, with
the standard deviation in units of the last decimal.}
\end{table*}

The external rows use a 256 resize with a 224 centre crop where our headline rows
use a single crop over the full field of view. Rescoring our own checkpoints under the
matched protocol costs 0.0025 to 0.0033 on all three architectures, so the
protocol is worth about a third of a point and the remaining gap to the upper
block is consistent with the upper block's far larger pretraining corpora,
though those rows also differ in architecture, objective and fine-tuning recipe,
so this is not an isolated cause. An earlier estimate of ours put it at about a point,
which was wrong by a factor of three.

\end{appendices}

\backmatter

\bmhead{Acknowledgements}

The author thanks Jin Lei (Tongji University) for supervision, for access to
computing resources, and for extensive discussion of the experimental design and
the statistical protocol. Claude Opus 5 (Anthropic) was used as a coding and
drafting assistant under the author's direction: it contributed to the
implementation of the training and evaluation code, to the drafting of this
manuscript, and to adversarial review of the claims. The author designed the
experiments, directed and ran them, and verified every performance number in the
tables and every value printed in the figures against the committed run
artifacts using dedicated verification scripts. A separate check holds the figure files shipped with this manuscript
byte for byte against the files their generators write, which guards a stale copy
and not the provenance of the generator's own inputs.

\section*{Declarations}

\textbf{Funding}\quad No funding was received for conducting this study.

\textbf{Competing interests}\quad The author declares no competing financial
interests or personal relationships that could have appeared to influence the
work reported in this paper.

\textbf{Ethics approval and consent to participate}\quad Not applicable. This
study uses only publicly released, de-identified datasets.

\textbf{Consent for publication}\quad Not applicable.

\textbf{Data availability}\quad ChestX-ray14 is publicly available from the
National Institutes of Health. CheXpert is available from Stanford under a use
agreement that permits non-commercial, non-clinical research use and prohibits
redistribution; the image files are therefore not redistributed.

\textbf{Materials availability}\quad Not applicable.

\textbf{Author contribution}\quad S.Q.\ designed the experiments, directed and ran
them, analysed the results, and wrote the manuscript.

\bibliography{lumen}

\end{document}